\documentclass[10pt]{article} %
\usepackage[accepted]{tmlr}

\usepackage{hyperref}
\usepackage{url}
\usepackage{amsthm}
\usepackage{bm}
\usepackage{soul}
\usepackage[]{xcolor}
\definecolor{rust}{rgb}{0.72, 0.25, 0.05}
\definecolor{princetonorange}{rgb}{1.0, 0.56, 0.0}

\definecolor{ann_color}{HTML}{42e6f5}

\usepackage{cancel}
\RequirePackage{graphicx}
\usepackage{booktabs}
\usepackage{caption}
\usepackage{subcaption}
\usepackage{lscape}
\usepackage{todonotes}
\usepackage[linesnumbered,ruled,vlined]{algorithm2e}
\providecommand{\rav}[1]{{#1}}

\usepackage{amsmath,amsfonts,bm}
\usepackage{mathabx}
\usepackage{dsfont}
\usepackage{nicefrac}

\makeatletter
\newcommand{\pushright}[1]{\ifmeasuring@#1\else\omit\hfill$\displaystyle#1$\fi\ignorespaces}
\newcommand{\pushleft}[1]{\ifmeasuring@#1\else\omit$\displaystyle#1$\hfill\fi\ignorespaces}
\makeatother

\newtheorem{theorem}{Theorem}[section]

\theoremstyle{remark}

\newcommand{\tuple}[1]{\ensuremath{\left\langle #1 \right\rangle}}

\newcommand{\fun}[1]{\ensuremath{\mathopen{}\mathclose\bgroup\left(#1\aftergroup\egroup\right)}}

\newcommand{\diracimpulsesymbol}{\ensuremath{\delta}}
\newcommand{\diracimpulse}[2]{\ensuremath{\diracimpulsesymbol_{#2}\fun{#1}}}

\newcommand{\states}{\ensuremath{\mathcal{S}}}
\newcommand{\actions}{\ensuremath{\mathcal{U}}}

\newcommand{\probtransitions}{\ensuremath{\mathbf{P}}} %
\newcommand{\rewards}{\ensuremath{\mathcal{R}}}

\newcommand{\state}{\ensuremath{s}}

\newcommand{\action}{\ensuremath{u}}
\newcommand{\jointaction}{\ensuremath{\bm{\action}}}

\newcommand{\act}[1]{\ensuremath{\mathit{Act}\ifthenelse{\equal{#1}{}}{}{(#1)}}}

\newcommand{\policy}{\ensuremath{\pi}}
\newcommand{\jointpolicy}{\ensuremath{\bm{\policy}}}

\newcommand{\valuessymbol}[2]{\ensuremath{{V}_{#1}^{#2}}}

\newcommand{\qvaluessymbol}[2]{\ensuremath{Q_{#1}^{#2}}}

\newcommand{\qvaluesproxysymbol}[2]{\ensuremath{\Tilde{Q}_{#1}^{#2}}}
\newcommand{\advantagessymbol}[2]{\ensuremath{A_{#1}^{#2}}}

\newcommand{\stationary}[1]{\ensuremath{\xi_{#1}}}

\newcommand{\pomdp}{\ensuremath{\mathcal{P}}}

\newcommand{\observationfn}{\ensuremath{O}}
\newcommand{\observation}{\ensuremath{o}}
\newcommand{\jointobservation}{\ensuremath{\bm{\observation}}}

\newcommand{\history}{\ensuremath{\tau}}
\newcommand{\jointhistory}{\ensuremath{\bm{\history}}}

 \newcommand{\encoderparameter}{\ensuremath{}}

\newcommand{\expectedsymbol}[1]{\ensuremath{\mathop{\mathbb{E}}\ifthenelse{\equal{#1}{}}{}{_{#1}}}}
\newcommand{\expected}[2]{\ensuremath{\expectedsymbol{#1} \left[ #2 \right]}}

\newcommand{\normal}[3]{\ensuremath{\displaystyle \ifthenelse{\equal{#3}{}}{\mathcal{N}(#1, #2)}{\mathcal{N}(#3\,;\, #1, #2)}}}

\newcommand{\overbar}[1]{\mkern 1.5mu\overline{\mkern-1.5mu#1\mkern-1.5mu}\mkern 1.5mu}
\newcommand{\overbarit}[1]{\,\overline{\!{#1}}}
\newcommand{\embed}{\ensuremath{\phi}}

\newcommand{\latentprobtransitions}{\ensuremath{\overbar{\probtransitions}}}

\newcommand{\latentrewards}{\ensuremath{\overbarit{\rewards}}}

\newcommand{\latentbeliefupdate}{\ensuremath{\overbar{\tau}}}

\newcommand{\localtransitionloss}[1]{L_{\probtransitions}}
\newcommand{\localrewardloss}[1]{L_{\rewards}}
\newcommand{\observationloss}[1]{\ensuremath{L_{\observationfn}}}
\newcommand{\beliefloss}[1]{\ensuremath{L_{\latentbeliefupdate}}}
\newcommand{\onpolicyrewardloss}[1]{\ensuremath{L_{\latentrewards}^{\varphi}}}
\newcommand{\onpolicytransitionloss}[1]{\ensuremath{L_{\latentprobtransitions}^{\varphi}}}

\newcommand{\KR}[1]{\ensuremath{\ifthenelse{\equal{#1}{}}{K_{\latentrewards}}{K_{\latentrewards}^{#1}}}}
\newcommand{\KP}[1]{\ensuremath{\ifthenelse{\equal{#1}{}}{K_{\latentprobtransitions}}{K_{\latentprobtransitions}^{#1}}}}

\newcommand{\originaltolatentstationary}[1]{{\latentprobtransitions_{\embed_{\encoderparameter}\stationary{\ifthenelse{\equal{#1}{}}{\policy}{#1}}}}}

\def\1{\bm{1}}

\DeclareMathAlphabet{\mathsfit}{\encodingdefault}{\sfdefault}{m}{sl}
\SetMathAlphabet{\mathsfit}{bold}{\encodingdefault}{\sfdefault}{bx}{n}

\newcommand{\abs}[1]{\ensuremath{\left| #1 \right|}}

\title{Local Advantage Networks \\for Multi-Agent Reinforcement Learning in Dec-POMDPs}

\author{\name Rapha\"el Avalos \email raphael.avalos@vub.be \\
      \addr Vrije Universiteit Brussel
      \AND
      \name Mathieu Reymond \email mathieu.reymond@vub.be \\
      \addr Vrije Universiteit Brussel
      \AND
      \name Ann Now\'e \email ann.nowe@vub.be\\
      \addr Vrije Universiteit Brussel
      \AND
      \name Diederik M. Roijers \email diederik.roijers@vub.be\\
      \addr Vrije Universiteit Brussel \\ City of Amsterdam
      }

\def\month{10}  %
\def\year{2023} %
\def\openreview{\url{https://openreview.net/forum?id=adpKzWQunW}} %

\begin{document}

\maketitle

\begin{abstract}
Many recent successful off-policy multi-agent reinforcement learning (MARL) algorithms for cooperative partially observable environments focus on finding factorized value functions, leading to convoluted network structures. Building on the structure of independent Q-learners, our LAN algorithm takes a radically different approach, leveraging a dueling architecture to learn for each agent a  decentralized best-response policies via individual advantage functions. The learning is stabilized by a centralized critic whose primary objective is to reduce the moving target problem of the individual advantages. The critic, whose network's size is independent of the number of agents, is cast aside after learning. Evaluation on the StarCraft II multi-agent challenge benchmark shows that LAN reaches state-of-the-art performance and is highly scalable with respect to the number of agents, opening up a promising alternative direction for MARL research.
\end{abstract}

\section{Introduction}

\emph{Reinforcement learning (RL)} \citep{Sutton1998ReinforcementIntroduction} is the branch of machine learning dedicated to learning through trial-and-evaluation by interaction between an agent and an environment. Research in RL has successfully managed to exceed human performance in many tasks including Atari games \citep{Mnih2015Human-levelLearning} and the challenging game of Go \citep{Silver2016MasteringSearch}. 

While single-agent RL has been highly successful, many real world tasks -- sensor networks \citep{Mihaylov2010DecentralizedNetworks}, wildlife protection \citep{Xu2020StayVersion}, and space debris cleaning \citep{Klima2018SpaceAnarchy} -- require multiple agents. When these agents need to act on local observations, or the problem is too large to centralize due to the exponential growth of the joint action space in the number of agents, an explicitly multi-agent approach is required. As such, \emph{Multi-Agent Reinforcement Learning (MARL)}  \citep{Busoniu2008ALearning,Hernandez-Leal2019ALearning, Shoham2007IfQuestion} introduces additional layers of complexity over single-agent RL.

In this paper, we focus on partially observable cooperative MARL where the agents optimize the same team reward. This setting introduces two main challenges that do not exist in single-agent RL. 
1) The \emph{moving target problem} \citep{Tuyls2012MultiagentProspects}: the presence of multiple learners in an environment makes it impossible for an agent to infer the conditional probability of future states. This invalidates most single-agent approaches, as the Markovian property no longer holds. 
2) The \emph{multi-agent credit assignment problem}: to learn a policy each agent needs to determine which actions contribute to obtaining the maximum reward. While in single agent RL this problem is only temporal, as the reward can be sparse and delayed, the shared reward increases the complexity of this problem as the agents also need to determine their individual contribution. 

\emph{Centralized Training with Decentralized Execution} (CTDE) \citep{Oliehoek2008OptimalPOMDPs, Foerster2018CounterfactualGradients, Lowe2017Multi-agentEnvironments}, has become a popular learning paradigm for MARL. The core idea behind CTDE is that even though decentralized execution is required the learning is allowed to be centralized. Specifically, during training, it is often possible to access the global state of the environment, the observations and actions of all agents allowing to break partial observability, which mitigates both the moving target problem and the credit assignment problem.

Most of the research in off-policy CTDE MARL for collaborative partially observable environments focuses on factorizing the joint Q-Value into local agent utilities such as QMIX \citep{Rashid2018QMIX:Learning} and QPLEX \citep{Wang2021QPLEX:Q-Learning}.

In this paper, we take a radically different approach. Our \emph{Local Advantage Networks (LAN)} algorithm learns for every agent the advantage of the best response policy to the other agents' polices. These local advantages, which are solely conditioned on the agent observation-action history, are sufficient to build a decentralized policy. In this sense, the architecture of LAN resembles independent Q-learners more than other CTDE approaches such as QMIX or QPLEX.
A key element of our solution is to derive a proxy of the local Q-value that leverages CTDE to stabilize the learning of the local advantages. For each agent the Q-value proxy is composed of the sum of the local advantage with the centralized value of the joint policy. 
Compared to the local Q-value, LAN's proxy is able mitigate the moving target problem, by integrating the changes of the other agents' policies faster%
, and to reduce the multi-agent credit assignment, by learning the local advantage function for each agent.
LAN is also highly scalable as the centralized value network reuses the hidden states of the local advantages to represent the joint observation-action history and the number of parameters of the centralized value does not depend on the number of agents. 
Finally, compared to QMIX and QPLEX which factorize the joint Q-value into individual utilities, LAN learns individual best-response Q-value proxies. %
This allows LAN to not have any restriction on the family of decentralized functions that it can represent, as opposed to QMIX. Indeed, in cooperative environments the optimal policies are best response policies.

We empirically evaluate LAN against independent Q-Learners \citep{Tan1993Multi-AgentAgents, Tampuu2015MultiagentLearning} and state-of-the-art algorithms for deep MARL, i.e., VDN \citep{Sunehag2018Value-decompositionReward}, QMIX and QPLEX, on the Starcraft Multi-agent Challenge (SMAC) benchmark \citep{Samvelyan2019TheChallenge}.
We show that on the 14 maps that compose the benchmark, LAN reaches similar performance of the SOTA in 11, surpasses the others algorithms with a large margin in 2, and under-performs in 1. In the maps with the most agents, LAN's centralized network uses up to $7$ times fewer parameters than QPLEX demonstrating the scalability of our algorithm. Furthermore, in two super hard maps, LAN learns a complex strategy based on an agent sacrificing itself to lure the enemies far from its teammates, showcasing LAN's capacity to mitigate the temporally extended multi-agent credit assignment problem. This strategy allows LAN to obtain a success rate of respectively $40\%$ and $90\%$ on two maps where the current state-of-the-art -- QPLEX -- struggles to obtain any wins. By improving performance on these two maps, LAN was able to achieve an average final performance on all 14 maps that is $10\%$ better than QPLEX's score.

Importantly, the objective of this new method is not to improve performance over the SOTA but rather to present an alternative research direction to factorizing the joint Q-value. 

\section{Background}
The setting considered in this paper are Dec-POMDPs \citep{Oliehoek2016APOMDPs, Oliehoek2008OptimalPOMDPs} $G = \langle \mathcal{A},\mathcal{S},\bm{\mathcal{U}},P,R,\bm{\mathcal{O}},O,\gamma \rangle$. At each time-step, every agent $a \in \mathcal{A}$ selects an action $u_a \in \mathcal{U}_a$ to form the joint action $\textbf{u} \in \bm{\mathcal{U}}$, where $\bm{\mathcal{U}} = \bigtimes_a \mathcal{U}_a$, that is processed by the environment to produce: a unique reward $r$ common to all agents; the next state $s' \in \mathcal{S}$; and the agents' joint observation $\bm{o} \in \bm{\mathcal{O}}$, where $\bm{\mathcal{O}} = \bigtimes \mathcal{O}_a$, with $o_a \in \mathcal{O}_a$ the observation of agent a. 
$\gamma \in [0,1)$ is the discount factor.
As the agents cannot access the real state of the environment they condition their policy on their observation-action history $\tau_a \in \mathcal{T}_a \equiv  (\mathcal{O}_a, \mathcal{U}_a)^*$, with $\bm{\tau} \in \bm{\mathcal{T}}$, where $ \bm{\mathcal{T}} = \bigtimes_a \mathcal{T}_a$ being the joint observation-action history. We refer to the observation-action history of an agent as its history, and the joint observation-action history as the joint history. To simplify the notations in this paper we assume that the observation function is deterministic. However the extension to stochastic observations is straightforward. With that setting, the next joint history $\bm{\tau}'$ is defined entirely by the current joint history, the joint action and the state $\langle \bm{\tau}, \bm{u}, s'\rangle$. The value, Q-value and advantage functions of the joint policy $\bm{\pi}$, which can be centralized or decentralized, are defined as: %

\begin{align*}
    V^{\bm{\pi}}(s, \bm{\tau}) &= \sum_{\bm{u}} \bm{\pi}(\bm{u} | \bm{\tau}) \big[ R(s, \bm{u}) + \gamma \sum_{s'} P(s' | s, \bm{u}) V^{\bm{\pi}}(s', \bm{\tau}') \big]
\end{align*}
\begin{align*}
    Q^{\bm{\pi}}(s, \bm{\tau}, \bm{u}) = R(s, \bm{u}) &+ \gamma \sum_{s'} P(s' | s, \bm{u}) V^{\bm{\pi}}(s', \bm{\tau}') \quad
    A^{\bm{\pi}}(s, \bm{\tau}, \bm{u}) = Q^{\bm{\pi}}(s, \bm{\tau}, \bm{u}) -  V^{\bm{\pi}}(s, \bm{\tau})
\end{align*}

We note that, if there is only a single agent a Dec-POMDP is a POMDP, and if this agent can observe the full state $s$ the POMDP is an MDP. 

DQN \citep{Mnih2013PlayingLearning} is a popular algorithm for MDPs 
that learns an approximation of $Q^*=\max_\pi Q^\pi$ with a neural network parametrized by $\theta$. This $\theta$ is learned through gradient descent by minimizing \mbox{$Q(s,u \mid \theta) - y^{DQN})^2$} with $y^{DQN}=r + \gamma \max_{u'} Q(s',u' \mid \theta)$. DQN uses a replay buffer to improve sample efficiency and to stabilize the learning. 
Dueling DQN \citep{Wang2016DuelingLearning} is a variant of DQN that learns both the value and the advantage, to then produce the Q-value as the sum of both instead of learning Q directly. This alternative architecture is motivated by the fact that having one part of the neural network that learns the general value of the state, and a second part that learns the effects of the actions - represented by the advantage - can be easier than learning both in the same network. DRQN uses a Recurrent Neural Network (RNN), such as a Gated Recurrent Network (GRU) \citep{Cho2014LearningTranslation} or an LSTM \citep{Hochreiter1997LongMemory}, to extend DQN to partial observablity (POMDP).%

\section{Related work}

Applying single agent RL algorithms to Dec-POMDPs, such as Independent Q-Learners (IQL) and Independent Actor-Critic, results in poor performance due to the moving target and multi-agent credit assignment problems \citep{Tan1993Multi-AgentAgents, Tampuu2015MultiagentLearning, Foerster2018CounterfactualGradients} -- with the exception of stateless normal form games \citep{Nowe2012GameLearning}. The replay buffer, fundamental to DQN, worsens the moving target problem as the sampled transitions are quickly outdated and off-environment as the policies evolve. Indeed, as all the agents are learning, states of transitions saved in the replay buffer might no longer be achievable by changing the policy of one agent. As removing the replay buffer does not lead to good polices, alternatives such as importance sampling and the use of fingerprints have been explored leading to small improvements \citep{Foerster2017StabilisingLearning}. \rav{In contrast, LAN's centralized value function mitigates the moving target problem sufficiently, which enables it to take advantage of the replay buffer and to reach state-of-the-art performance. }

COMA \citep{Foerster2018CounterfactualGradients} and MADDPG \citep{Lowe2017Multi-agentEnvironments} introduced CTDE to Deep MARL by building on single-agent actor-critic algorithms but replacing the local critic with a centralized one to improve the quality of the value estimation guiding the updates. In comparison, our method, LAN, is a value-based algorithm making it more sample-efficient. While LAN's joint value is also a centralized critic, it plays an intrinsically different role, as it fosters learning coordination between the local advantage functions.

Centralized Q-Learning (CQL) and Independent Q-Learners (IQL) form the two extremes of value-based methods for MARL. On the one hand, CQL learns a unique Q-Value conditioned on the full joint action space and the joint history. While in this setting the optimal performance is better or equal to the decentralized one due to the reduction of partial observability, the agents are no longer autonomous as they rely on a central entity for execution. In addition, this algorithm does not scale well due to the exponential increase of the joint action space in the number of agents. On the other hand, IQL learns for each agent a local Q-Value conditioned on its local observation-action history. This algorithm is heavily affected by the moving target problem. However, in settings with limited interactions between agents, the moving target problem is not as intense and IQL can show good performance. 

\begin{figure*}
    \centering
    \includegraphics[width=\textwidth]{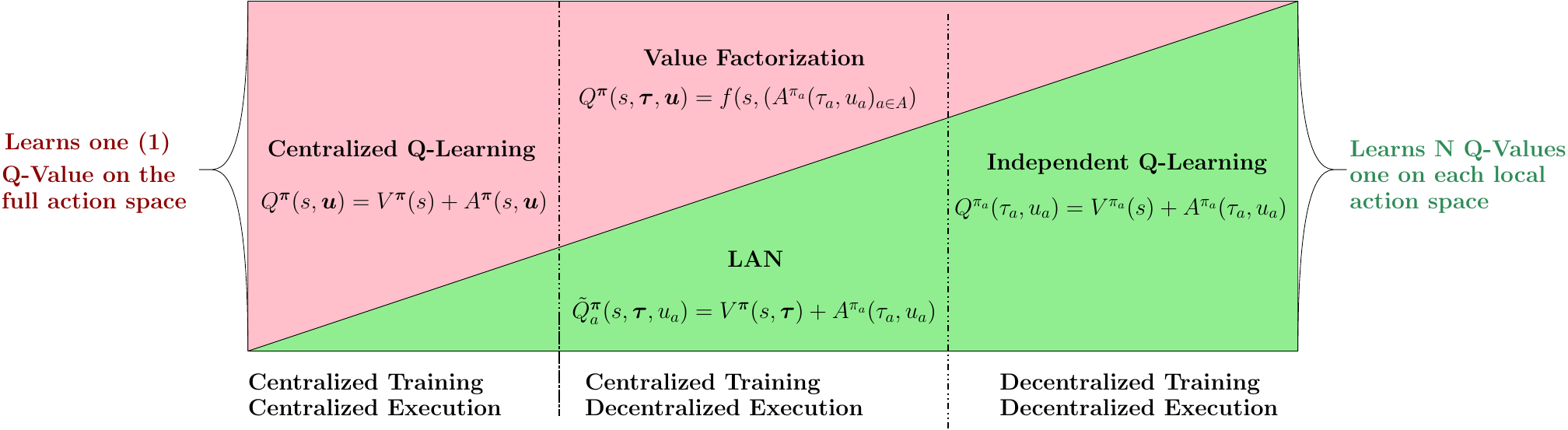}
    \caption{Comparison diagram between Centralized Q-Learning, Independent Q-Learning, Value factorization methods and LAN.}
    \label{fig:compare}
\end{figure*}

Value factorization (VDN, QMIX, QPLEX) emerged as the main alternative in recent years. Closer to CQL than IQL, those algorithms learn a unique Q-Value over the joint action space. Its factorized architecture allows recovering for each agent a utility function for action selection. To ensure that the agents select the same action during training with the centralized component and during decentralized execution, the factorization follows the individual global max (IGM) principle: the maximizing joint action of the joint Q-value must be equal to the joint action that results from maximizing the local utilities. The factorization usually enforces a monotonicity constraint to ensure IGM, i.e., for each agent the derivative of the joint Q-value to the agent's local utility is positive. VDN is the first algorithm of this kind and decomposes the joint Q-value into a simple sum. QMIX extends VDN by learning state-dependent positive weights. The state dependency broadens the family of Q-value functions that can be learned, and the positive weights constraint ensures IGM. While QMIX achieves good performance and improves over VDN, the monotonicity constraints still limits the family of functions learnable. QATTEN \citep{Yang2020Qatten:Learning} extends QMIX by using multi-head attention \citep{Vaswani2017AttentionNeed} to compute the mixing weights. More recently, QPLEX extends QATTEN by transferring the IGM principle from the Q-value to the advantage function. At the cost of twice as many parameters on average and a more complex mixing network, QPLEX outperforms QMIX on SMAC. 
In contrast to those algorithms, LAN does not factorize the joint Q-value into individual agents utilities but learns an individual Q-value proxy for every agent. This result in LAN's architecture being able to represent any decentralized policy, as opposed to QMIX and VDN.

Figure~\ref{fig:compare} presents a visual comparison of the structural differences between CQL, IQL, value factorization and LAN. This figure highlights the fact that while value factorization and LAN are both CTDE methods, LAN is closer to IQL as it learns for each agent a Q-value on its local action space. 

Improving multi-agent exploration or scalability regarding the action space in Dec-POMDPs have been successfully explored by MAVEN \citep{Mahajan2019MAVEN:Exploration} and RODE \citep{Wang2020ROMA:Roles}. Both works are orthogonal to ours, and while they use QMIX as a base algorithm they could also be applied to LAN. For this reason we do not include them as baselines.

Recently, MAPPO \citep{Yu2021TheGames} and IPPO \cite{deWitt2020IsChallenge} showed that actor-critic-based algorithms could achieve good performance on cooperative MARL. However, they require significantly more interactions, 10 million timesteps instead of 2 million, and more computing power. 
Comparison with those two algorithms is also harder because MAPPO changed the state space, IPPO changed the difficulty of the enemy team, and they do not use the same version of the environment. 
Also, they both have different hyperparameters per map whereas the other algorithms have one set of hyperparameters for the full benchmark challenge. 
However, just like LAN, both MAPPO and IPPO propose an alternative to off-policy value factorization. While comparing the three methods might not be straightforward due to the need to retune the three algorithms on a fixed version of SMAC, a further comparative study would help to better understand the strengths and weaknesses of each algorithm.

\section{Method}
\label{sec:method}

In this section, we present \textbf{Local Advantage Networks (LAN)}, a novel value-based algorithm for collaborative partially observable MARL. LAN goes in the opposite direction of the current state-of-the-art in MARL, which focuses on factorizing the Q-value of the joint policy $Q^{\bm{\pi}}$ into individual utilities. Instead, LAN learns for each agent the advantage of the best response policy to the other agents' policies. The local advantages are only conditioned on the own agent's history allowing for decentralized execution. The main contribution of LAN is to stabilize the learning of those advantages by leveraging CTDE to use the value of the joint policy $V^{\bm{\pi}}$ to coordinate their learning. The centralized nature of $V^{\bm{\pi}}$ allows to reduce the partial observability, mitigate the moving target problem and the multi-agent credit assignment problem. By combining the local advantages with the centralized value, LAN derives a proxy of the individual Q-value of each agent and can simultaneously learn all components with DQN. Two key differences with a factorized Q-function are: (1) that LAN does not learn the Q-value of the joint policy, which is in fact more difficult to learn than the value $V$, and its factorization, but proxies of the individual Q-Values and (2) that in contrast to VDN and QMIX, LAN's architecture does not limit the the family of decentralized policies it can represent. We note that QPLEX can also represent all these policies at the cost of a more complex architecture.%

\paragraph{Best response policies}
We start from the observation that in a Dec-POMDP when the agents reach an optimal policy, their individual policies are best responses to the other agents' policies. Indeed, if one agent could improve its policy while the other agents polices are fixed, the joint policy cannot be optimal as the agents share the same reward. 
Based on this observation, LAN focuses on learning best response polices.

To better understand how to learn best response policies, we first focus on a single agent $a \in \mathcal{A}$ and assume that the joint policy of the other agents $\bm{\pi}_{-a}$ is fixed. As in \citep{Foerster2017StabilisingLearning}, we derive from the Dec-POMDP $G$ a POMDP $G_a=\langle\Tilde{\mathcal{S}}, \mathcal{U}_a, P_{a}, \mathcal{O}_a, O_a, R_{a}, \gamma\rangle$, with $\Tilde{S} = \langle \mathcal{S}, \bm{\mathcal{T}}_{-a} \rangle$ being the original state space extended with the observation-action histories of the other agents, $P_a$ and $R_a$ are defined as follows:
\begin{align*}
    P_a(\langle s', \bm{\tau}'_{-a} \rangle |\langle s, \bm{\tau}_{-a} \rangle , u_a) &= \sum_{\bm{u}_{-a}} \bm{\pi}_{-a}(\bm{u}_{-a}|\bm{\tau}_{-a}) P(s'| s, \langle u_a, \bm{u}_{-a} \rangle) p(\bm{\tau}_{-a}'| \bm{\tau}_{-a}, s, s', \bm{u}_{-a}) \\
    R_a(\langle s, \bm{\tau}_{-a}, u_a) &= \sum_{\bm{u}_{-a}} \bm{\pi}_{-a}(\bm{u}_{-a} | \bm{\tau_{-a}}) R(s, \langle u_a, \bm{u}_{-a} \rangle)
\end{align*}

The value, Q-value and advantage of $G_a$ can then be derived as follows, with $p(\Tilde{s}|\tau_a)$ the probability of being in an extended state $\Tilde{s} \in \Tilde{\mathcal{S}}$ when $\tau_a$ is agent $a$'s local history.

\begin{align*}
    \begin{split}
        V^{\pi_a}(\tau_a) = \sum_{u_a} \pi_a(u_a|\tau_a) \sum_{\Tilde{s}} p(\Tilde{s}|\tau_a)\sum_{\bm{u}_{-a}} \bm{\pi}_{-a}(\bm{u}_{-a}|\bm{\tau}_{-a})
     \big[ R(s, (u_a, \bm{u}_{-a})) + \gamma \sum_{s'} P(s'|s, \langle u_a, \bm{u}_{-a} \rangle) V^{\pi_a}(\tau_a')\big]
    \end{split}\\
    \begin{split}
        Q^{\pi_a}(\tau_a, u_a) = \sum_{\Tilde{s}} p(\Tilde{s}|\tau_a) \sum_{\bm{u}_{-a}} \bm{\pi}_{-a}(\bm{u}_{-a}|\bm{\tau}_{-a})
    \big[ R(s, (u_a, \bm{u}_{-a})) + \gamma \sum_{s'} P(s'|s, \langle u_a, \bm{u}_{-a} \rangle) V^{\pi_a}(\tau_a')\big]
    \end{split}\\
    \begin{split}
        Q^{\pi_a}(\tau_a, u_a) = V^{\pi_a}(\tau_a) + A^{\pi_a}(\tau_a, u_a)
    \end{split}
\end{align*}

\paragraph{Partial observability} Due to the partial observability, agent $a$ needs to disambiguate the state of $G_a$ corresponding to the original state $s$ and the joint history of the other agents $\bm{\tau}_{-a}$. As the environment is no longer Markovian, the agent needs to base its policy on a belief over the extended state. The most straightforward way to compute this belief is to keep the full history of the agent. However, this strategy does not scale well in the number of time-steps or state space. As analyzed in the work on influence-based abstractions \citep{Oliehoek2012Influence-basedSystems}, in a Dec-POMDP maintaining a belief over the subset of features that allows to locally regain the Markovian property is sufficient, using the property of d-separation. This belief is much more compact than keeping track of the entire action-observation history, and therefore offers the possibility to keep a fully sufficient representation that remains tractable. 
In the ideal case, the RNN's history representation will capture the belief over the d-separating features, enabling the reinforcement learning agent to learn an optimal Dec-POMDP policy. In practice of course, we aim to closely approximate such a representation, but are often uncertain of its existence, or of its size if it does exist. 

Applying DQN to the single-agent POMDP $G_a$ learns, for each agent $a$, the best response policy to $\bm{\pi}_{-a}$, as the probability distribution over the relevant features $P_a$ results from executing fixed policies for the other agents.
A naive solution to learn good decentralized policies would therefore be to improve each agent successively. However, this approach fails if the environment requires the agents to explore simultaneously to find the optimal policy. On the other hand, optimizing $Q^{\pi_a}$ for all the agents simultaneously, i.e., Independent Q-Learning (IQL) \citep{Tan1993Multi-AgentAgents, Tampuu2015MultiagentLearning} also has key downsides. While IQL allows agents to explore together, it does not perform well in more complicated tasks due to the moving target problem as it ignores that the environment $G_a$ perceived by agent $a$ is shifting as $\bm{\pi}_{-a}$ evolves. 
So while we need agents that learn together, they need to do so in a coordinated manner. 

\paragraph{Q-Value proxy} 
LAN simultaneously learns best response policies and mitigates the moving target problem. These best response policies are expressed as \emph{local advantage functions} that are solely conditioned on the agent's observation-action history, $A^{\pi_a}(\tau_a, u_a)$, allowing for decentralized execution. To coordinate the learning of those local advantage functions, following the CTDE paradigm, LAN leverages full information about the state and the other agents observation-action history at training time via a centralized value function $V^{\bm{\pi}}$. More specifically, LAN derives $\Tilde{Q}^{\bm{\pi}}_a$ a proxy of the local Q-value $Q^{\pi_a}$ for each agent $a \in A$.
\begin{align}
    \Tilde{Q}^{\bm{\pi}}_a(s, \bm{\tau}, u_a) = V^{\bm{\pi}}(s, \bm{\tau}) + A^{\pi_a}(\tau_a, u_a)
    \label{eq:q_tilde}
\end{align}

The proxy is constructed by summing the local advantage $A^{\pi_a}$ with the centralized value of the joint policy $V^{\bm{\pi}}$. While $\Tilde{Q}^{\bm{\pi}}_a$ is not a real Q-value and it is conditioned on the full state and the joint history $\bm{\tau}$ it can be used to extract decentralized policies as the maximizing actions only depend on the agent's history $\tau_a$, as shown by equation \ref{eq:adv_action}. We obtain this equation by remarking that for both decomposition of $Q^{\pi_a}$ and $\Tilde{Q}^{\pi}_a$, the local and centralized values are not conditioned by the agent's actions.
\begin{align}
    \arg\max_{u_a} \Tilde{Q}^{\bm{\pi}}_a(s, \bm{\tau}, u_a) &= \arg\max_{u_a} A^{\pi_a}(\tau_a, u_a) = \arg\max_{u_a} Q^{\pi_a}(\tau_a, u_a)
    \label{eq:adv_action}
\end{align}

LAN uses DQN to learn the individual Q-value proxy $\Tilde{Q}^{\bm{\pi}}_a$ for all agents $a\in A$ simultaneously. This allows LAN to learn the local advantages $A^{\pi_a}$ and the centralized value $V^{\bm{\pi}}$ in parallel by optimizing a unique loss, resulting in an efficient learning scheme.
LAN's DQN target for agent $a$ is defined as follows  with the subscript $t$ referring to a delayed copy of the networks to increase learning stability \citep{vanHasselt2015DeepQ-learning}. Appendix \ref{sec:alg} contains the pseudo code of LAN.
\begin{align}
    y_a &= r + \gamma \Tilde{Q}^{\bm{\pi}}_{t_a}(s', \bm{\tau}', \arg\max_{u_a'} \Tilde{Q}^{\bm{\pi}}_{a}(s', \bm{\tau}', u_a')) = r + \gamma [V^{\bm{\pi}}_t(s', \bm{\tau}') + A^{\pi_a}_t(\tau_a', \arg\max_{u_a'} A^{\pi_a}(\tau_a', u_a'))]
    \label{eq:target}
\end{align}

The following Theorem, shows that our Q-value proxy is an unbiased estimator of the local Q-value it approximates.

\begin{theorem}
    For any agent $a \in \mathcal{A}$, and any realisable local history $\history_a \in \mathcal{T}_a$, and any action $\action_a \in \actions_a$
, the Q-value proxy $\tilde{Q}_a$ is an unbiased estimator of the local Q-value $Q^{\policy_a}$
\begin{align}
    \expectedsymbol{\state, \jointhistory_{-a} \sim p\fun{\cdot \mid \history_a}} \qvaluesproxysymbol{a}{}\fun{\state, \tuple{\jointhistory_{-a}, \history_{a}}, \action_a} = \qvaluessymbol{}{\policy_a}\fun{\history_a, \action_a}
\end{align}
\end{theorem}

We prove the Theorem in Appendix \ref{sec:proof}. In a nutshell, this Theorem shows that by optimizing the Q-value proxy we are optimizing in the same direction of the local Q-value.

\begin{figure*}
    \centering
    \includegraphics[width=1\textwidth]{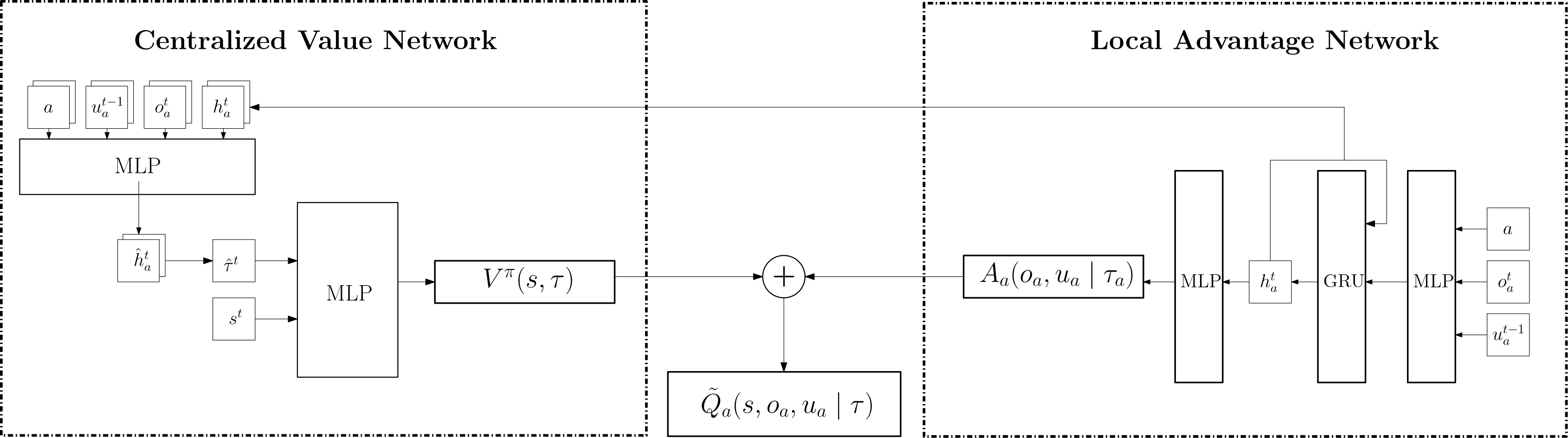}
    \caption{Architecture of LAN.}
    \label{fig:architecture}
\end{figure*}

Compared to the local Q-value $Q^{\pi_a}$, the learning of LAN's proxy $\Tilde{Q}^{\bm{\pi}}_a$ has two interesting properties that help stabilize and coordinate the learning, and give an intuition on how LAN solves the task as a whole. We note that these properties result from applying DQN to LAN's Q-value proxies to all agents in parallel, and cannot be tested independently.

\textbf{Property 1:} $\Tilde{Q}^{\bm{\pi}}_a$ mitigates the moving target problem, which results from
all the agents learning at the same time. This simultaneous learning allows the agent to explore together, which is necessary to find an optimal strategy in non-monotonic environments, but because of it the environment is constantly changing and locally loses its Markovian property. To provide meaningful updates and prevent the learning to plateau prematurely as in IQL, the updates need to reflect as closely as possible the ever changing environment. LAN achieves this thanks to the centralized value, which coordinates the learning of all the local advantages. This happens in two steps. First, as an update of $\Tilde{Q}^{\bm{\pi}}_a$ results in the update of both the centralized value and the local advantage with the same transitions, a modification of a local advantage function results in a change of the centralized value. Second, as the centralized value is part of the target update of every agent's Q-value (eq. \ref{eq:target}), the change is then propagated to the other agents' advantage.

\textbf{Property 2:} $\Tilde{Q}^{\bm{\pi}}_a$ mitigates the multi-agent credit assignment problem. As the centralized value function approximates the expected return of the joint policy, the agents can easily evaluate the effect of their actions on the effective return simply by subtracting it from the centralized value. This difference is learned by the local advantages. Indeed, by applying DQN to $\Tilde{Q}^{\bm{\pi}}_a$ the \emph{induced update} of the local advantage network of agent $a$ (eq. \ref{eq:credit}) is similar to the one used by COMA \citep{Foerster2018CounterfactualGradients} to reduce the multi-agent credit assignment problem. We stress the fact that we learn all the networks in parallel with the Equation~\ref{eq:target}.   
\begin{align}
    y_{A_a} = r + \gamma \Tilde{Q}^{\bm{\pi}}_{t_a}\left(s', \bm{\tau}', \arg\max_{u_a'} \Tilde{Q}^{\bm{\pi}}_{a}\left(s', \bm{\tau}', u_a'\right)\right) - V^{\bm{\pi}}\left(s, \bm{\tau}\right)
    \label{eq:credit}
\end{align}

Additionally, we also have two intuitions regarding LAN's performance. While we were not able to prove them, we believe that they are still valuable leads to explore.

\textbf{Intuition 1:} $\Tilde{Q}^{\bm{\pi}}_a$ allows to provide better update targets by breaking the partial observability. 
In a POMDP, the same observation-action history can be linked to different states forcing the agent to learn a Q-value that marginalizes over the possible states. In a Dec-POMDP this aspect is even more apparent as all the agents $a \in A$ need to marginalize over the possible states but also over the possible joint histories of the other agents $\langle s, \bm{\tau}_{-a} \rangle$ as shown by the derivation of $G_a$. By its conditioning on the next state and the joint history $\langle s', \bm{\tau}' \rangle$, LAN's DQN target does not suffer from the partial observability and can therefore provide updates taking into account this information. 
As highlighted by \citep{Lyu2021ContrastingLearning}, using a centralized target to learn a decentralized object might lead to high variance updates. The authors mention that the choice of a centralized versus decentralized critic is a bias-variance trade-off. In LAN, the value is centralized while the Advantage is decentralized. This means that LAN by using the Q-value proxy (not as precise as the real Q-values) to compute the targets induces a bias which in turn reduces the variance of the updates.%

\textbf{Intuition 2:} $\Tilde{Q}^{\bm{\pi}}_a$ reduces the learning complexity associated with decentralized policy optimization. Typically, extracting a policy from a value-based algorithm involves selecting the action that maximizes the Q-value or advantage, as they have the same action ordering. However, advantage and value functions exhibit different learning complexities, depending on the characteristics of the environment. While the advantage function learns the impact of each action on the overall return, the value function learns the expected cumulative return, necessitating more marginalization over different states and other agents' histories. This distinction motivated the introduction of Dueling DQN in MDPs \citep{Wang2016DuelingLearning}. Nonetheless, learning the advantage function in isolation is not feasible; it requires learning the corresponding value function, which suffers from both the partial observability and the moving target problem. Therefore, LAN's proxy provides a straightforward and efficient approach to learn local advantages without relying on local values.

\subsection*{Architecture}

To overcome the partial observability the local advantages networks use a GRU which learns to represent the observation-actions history into a hidden state $h_a$, with the aim to capture the necessary features to locally regain the Markov property as stated above. This hidden state is then used to compute the local advantages. LAN leverages the work done at the agent level to represent $\tau_a$ to build a representation of $\bm{\tau}$. 

For each agent $a$ the centralized value network combines the id $a$ of the agent with its hidden state $h_a$, its last observation $o_a$ and its last action $u_a$  into a vector $\Tilde{h}_a = [h_a, o_a, u_a, a]$. 
To represent $\bm{\tau}$ efficiently we first embed $\Tilde{h}_a$ into $\hat{h}_a$ for all agents with a shared network and sum those embedding. The embedding allows to limit the potential information loss of the summation, and this combination performs better than concatenation. Finally, the value is computed from $\bm{\tau}$ using an MLP.
LAN's architecture, represented in Figure \ref{fig:architecture}, provides two main benefits. First, the centralized value network does not learn a second recurrent network, which are knowingly difficult to train. Second, as the embedding for all agents are computed with the same weights, the number of parameters of the centralized value network does not depend on the number of agents. 

As the policies are deterministic, the local advantages should be negative with the maximizing value equal to $0$. However as \citep{Wang2016DuelingLearning} studies, even when computing the real Q-value in single agent MDP enforcing this constraint has a negative impact on the learning. Their experiments showed that applying the following transformation to the output of the neural network provides better stability.
\begin{align}
    A^{\pi_a}(\tau_a, u_a) \leftarrow A^{\pi_a}(\tau_a, u_a) - \frac{1}{|U_a|} \sum_{u \in u_a} A^{\pi_a}(\tau_a, u) 
\label{eq:adv_mean}
\end{align}
In the single agent case, this results in the learned advantage to differ from the real advantage by a fixed offset. In LAN, as the centralized value is shared between all the agents, enforcing the local advantages to have a zero mean means that the offset will be shared between all the agents. As in \citep{Wang2016DuelingLearning}, we investigated enforcing negative advantages and observed that the learning was also highly impacted by it in LAN. While sharing the offset between the agents can have a positive impact on collaboration it can also hinder the learning by adding an additional constraint on both networks. Appendix \ref{app:adv} reports LAN's performance with the mean constraint (eq. \ref{eq:adv_mean}). 
Therefore, in LAN we do not apply any constraint on the output of the advantage network.

\section{Experiments}

To benchmark LAN we use the StarCraft Multi-Agent Challenge\footnote{We use version SC2.4.6.2.69232 and not SC2.4.10. Performances are not comparable between versions.} (SMAC) \citep{Samvelyan2019TheChallenge}, a set of environments that runs in the popular video game StarCraft II. SMAC does not focus on the full game but rather on micromanagement tasks where two teams of agents - possibly heterogeneous and imbalanced - fight. A match is considered won if the other team is eliminated within the time limit. The time limits differ per task. Each agent only observes its surroundings and receives a team reward proportional to the damage done to the other team plus bonuses for killing an enemy and winning. The action space of each agent consists of a move action to each cardinal direction, a no-op action, and an attack action for each enemy which is replaced by a heal action for each team member for the Medivacs units. The attack/heal action only affects units within range. As the agent's observation and action space are linearly dependent on the number of agents to perform well scalability is a key issue. SMAC also provides the real state of the environment, which we use as input for the centralized value. The benchmark is composed of 14 different maps that are designed to assess different aspects of cooperation. They are ranked into 3 categories: easy, hard, and super hard maps.

\subsection{Configuration}

To ensure a fair comparison, the decentralized network architecture, the version of the game, the $\varepsilon$-annealing parameters, the batch size, the replay buffer size, the use of a single environment, and the use of a unique set of parameters across all maps is consistent with the QMIX and QPLEX papers. Appendix \ref{app:implem} lists the hyper-parameters used, and Appendix \ref{app:adv} \rav{reports the results of a variation of LAN where we force the advantage to have a zero-mean as in Dueling DQN \citep{Wang2016DuelingLearning}.} %
The training and evaluation follows the procedure described in \cite{Samvelyan2019TheChallenge}, namely $2$ million training timesteps, and evaluation of the decentralized greedy polices over $32$ episodes every $10k$ timesteps.
We train LAN on at least $10$ different random seeds and report the median of the battle win rate over the learning time as well as the first and third quantiles.

\begin{figure*}[t!]
    \centering
    \includegraphics[width=0.7\textwidth]{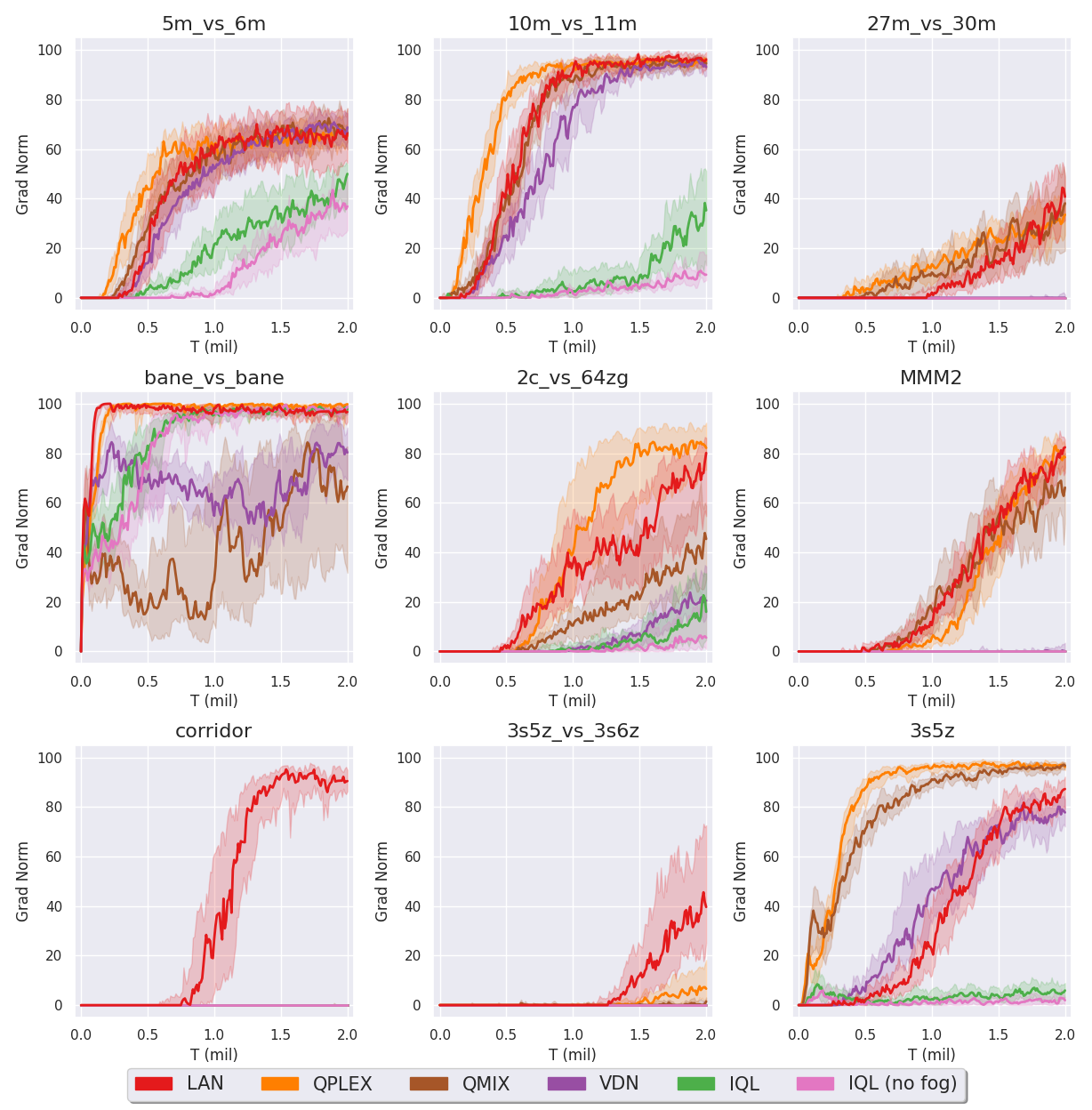}
    \caption{Median battle won rate during learning on 9 maps of SMAC. Each algorithm is run on at least 10 different seeds per map. Following the evaluation method of \cite{Samvelyan2019TheChallenge}, we train the agents on 2 million steps and plot the median, 1st and 3rd quantiles. IQL (no fog) is introduced in \ref{subseq:movingtarget}}
    \label{fig:maps}
\end{figure*}

\subsection{Results}

\begin{figure}[h!]
    \centering
    \includegraphics[width=0.65\textwidth]{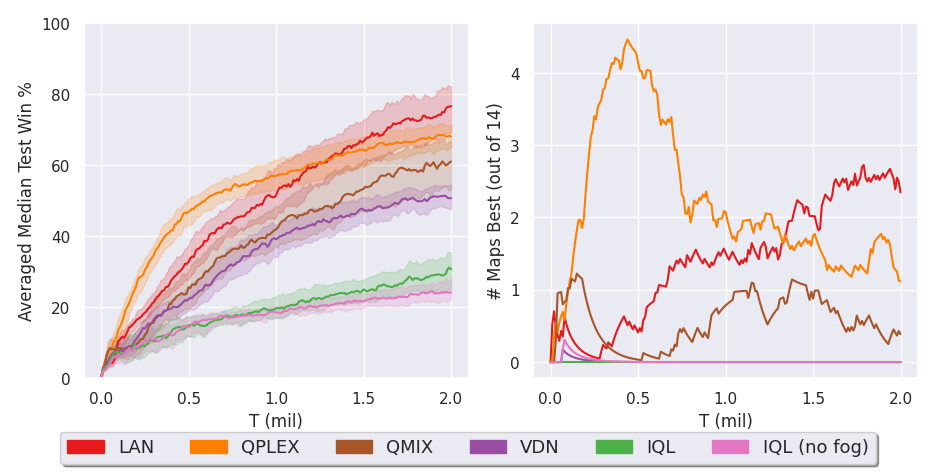}
    \caption{(Left) Averaged median test win on the 14 maps during learning. Shaded area denotes average first and third quantile. (Right) Number of maps where the algorithms are first by at least 1/32 during learning.}
    \label{fig:perf}
\end{figure}

We compare LAN to IQL, VDN, QMIX and QPLEX. For the first three algorithms we used the implementation of QMIX and we used the official implementation of QPLEX. %
In the following, we present LAN's performance on 9 maps (Figure \ref{fig:maps}). The other maps are presented in Appendix \ref{app:remaining}. The first row features fights between marines with an increasing number of agents and the enemy controlling more units. The second row is composed from left to right of a balanced map with 24 heterogeneous units per team, a map where 2 power-full units fight a swarm of 64 smaller enemies, and an unbalanced heterogeneous map with a medic units that as a side effect increases the action space. The last row shows the result on two super-hard maps where the baselines do not reach any wins, and a map where LAN seems to under-perform. Finally, we discuss LAN's average performance across all maps (Figure \ref{fig:perf}).

In the maps of the first row of Figure \ref{fig:maps}, two unbalanced teams with homogeneous units fight against each other, with our team composed of fewer units than the enemy: in \texttt{5m\_vs\_6m} 5 agents fight 6 enemies, in \texttt{10m\_vs\_11m} 10 agents fight 11 enemies, and in \texttt{27m\_vs\_30m} 27 agents fight 30 enemies. The ratio between the number of agents and the number of enemies makes the map \texttt{10m\_vs\_11m} easier compared to the other two. In the map \texttt{27m\_vs\_30m}, both the number of agents and the dimension of the observation and action space constitute a real challenge for MARL.
In those three maps, LAN dominates IQL and performs on par with SOTA. 
First, as IQL is a natural ablation of LAN, we deduce from this experiment that the centralized value introduced by LAN does indeed help to coordinate the learning of the agents and that LAN can address the shortcomings of IQL. 
Second, LAN does not only performs on par with the SOTA, and slighty outperforms the other algorithms in the more difficult map, it is also more scalable than QMIX and QPLEX in terms of parameters of its centralized component with respect to the number of agents (Table \ref{tab:param_table}). Indeed, between \texttt{5m\_vs\_6m} and \texttt{27m\_vs\_30m} the number of agents is multiplied by $5.4$ and the number of parameters of LAN's centralized value is only multiplied by a factor of $2$, while for the centralized component of QMIX and QPLEX this factor is respectively $8.8$ and $16.5$. 

The second row of Figure \ref{fig:maps}, is composed of two hard and one super-hard maps.
The first one, \texttt{bane\_vs\_bane}, opposes two large and balanced teams of 24 heterogeneous units. We observe that while IQL easily reaches 100\% of winning rate, VDN struggles to learn and QMIX fails to learn. This hints at a limitation of both monotonous mixing strategies regarding scaling to a large number of agents, supporting our claim that an alternative research direction to value factorization is needed. QPLEX is able to learn the perfect strategy at the cost of doubling the number of parameters compared to QMIX. LAN also learns to consequently eliminate the opposing team and reaches a perfect score with $5$ times fewer parameters than QPLEX. 
The second map, \texttt{2c\_vs\_64zg}, matches two powerful agents against 64 weaker agents. The numerous enemies make the action space very large, with $70$ actions, which is a known challenge in RL \citep{Zahavy2018LearnLearning}. In this map, QPLEX reaches a final performance of $83\%$ win rate followed closely by LAN with $80\%$, while QMIX, VDN and IQL score respectively around $50\%$, $20\%$ and $15\%$ win rate. %
The third map, \texttt{MMM2}, features two unbalanced heterogeneous teams, with the enemy team having 2 additional units, and is the only map including medical units. While IQL and VDN do not obtain any wins, QMIX and QPLEX score $60\%$ and $80\%$ respectively. LAN obtains the same final performance as QPLEX. 

\begin{table}[t]
    \caption{Number of parameters (x1000) of the value function in LAN vs.\ the mixing network in QPLEX/QMIX for the first 4 maps of Figure \ref{fig:maps}. See Appendix \ref{app:smac} for the other maps. The dependency of the dimension of the observation and action space in the number of agents is the only cause of the difference in the number of parameters of LAN's centralized value network in the different maps}
    \centering
    \begin{tabular}{ccccc}
        \toprule
         &  \textbf{5m\_vs\_6m} &  \textbf{10m\_vs\_11m} &  \textbf{27m\_vs\_30m} &  \textbf{bane\_vs\_bane}\\
        \midrule
        \textbf{LAN}     &   56 &     68 &    111 & 125 \\
        \textbf{QPLEX}   &   43 &    106 &    709 & 555 \\
        \textbf{QMIX}   &  32 &    70 &   283 & 241 \\
        \bottomrule
        \end{tabular}
    \label{tab:param_table}
\end{table}

The last row of Figure \ref{fig:maps} presents LAN's performance on 2 super-hard maps alongside the easier version of one of those maps. 
In the super hard map \texttt{corridor}, 6 agents of type 'zealot' fight a team 24 enemies of type 'zerlings'. While the SMAC paper claimed that the only solution for this map was to take advantage of the terrain (a spawning zone connected to a second zone by a corridor) to limit the number of enemies that can attack our agents, LAN discovered another solution. One agent lures part of the enemies to a remote location while the rest fights the remaining enemies. After killing the bait a fraction of the enemies attack our agents while the majority go through the corridor to reach the second zone. Our agents defeat their attackers, and after regenerating part of their shields move to the second zone to finish off the enemies.
While the current SOTA flattens to zero, LAN obtains an almost perfect score with around $90\%$ success rate.
On the next super hard map, \texttt{3s5z\_vs\_3s6z}, LAN learns good decentralized policies with a performance at around $40\%$. The only other algorithm that was able to achieve any wins is QPLEX with less than $10\%$. The strategy is similar as the one learned in corridor, a stalker (long-range unit) baits most of the enemy's zealots (close combat units) into targeting him. It then flees far away from his teammates and sacrifices himself so that the other agents can kill the stalkers and remaining zealots. The agents can then easily kill the remaining enemies as they are no longer protected by any long-range support. 
The last map of Figure \ref{fig:maps}, \texttt{3s5z}, is the balanced version of the previous map and therefore easier. In this map, LAN reaches $87\%$ median battle win rate, whereas VDN only scores $80\%$, and QMIX and QPLEX obtain $97\%$. This underperformance is intriguing as LAN performs better than the other algorithms in the harder version of this map. By visualizing the learned policies in \texttt{3s5z} we discovered that LAN converges to two different policies: a) a basic confrontation policy which is the policy learned by QMIX and QPLEX; b) a baiting strategy identical to the one learned in \texttt{3s5z\_vs\_3s6z}. We also remark that LAN appears to still be learning and might converge to the same performance as the other QPLEX if given more time. 

LAN's performance in last two super-hard maps can be attributed to is its ability to train an agent to lure the enemies and to sacrifice itself for the team's survival. We believe that this behavior is easier to discover with LAN than with the mixing algorithms because of the shared Value network, as it allows dead agents to benefit directly from the rewards scored by the other agents after their death.
LAN, by focusing on learning best response policies instead of factorizing a joint Q-value, learns for each agent the policy that maximizes the team return.
On the other hand, QMIX and QPLEX introduce individual rewards through factorization, which agents learn to maximize. However, if these individual rewards do not align with the team reward, as is the case in baiting strategies, mixing algorithms struggle to learn effectively. 
The complex strategy learned by LAN demonstrates its capacity to mitigate effectively the multi-agent credit assignment problem.

As in the SMAC benchmark and QPLEX papers, Figure \ref{fig:perf} shows, on the left plot, LAN's general average performance on the 14 maps that composes the SMAC benchmark, and, on the right plot, the number of maps where each algorithm outperforms the others by a margin of at least $1/32^{\text{th}}$. IQL only achieves $30\%$ averaged median test wins and is the best on $0$ maps. This under-performance was expected as it is the only fully decentralized learning algorithm, and because it is highly vulnerable to the moving target problem. At the beginning of the learning, VDN and QMIX show similar performance, but QMIX takes the lead obtaining $60\%$ and beating VDN by $8\%$. QPLEX learns faster than the other algorithms and reaches the same final performance of QMIX in just a million timesteps to obtain $67\%$ at the end of the learning. Finally, LAN learns faster than the baselines except QPLEX, which it exceeds at around $1.25\times 10^6$ timesteps. LAN finishes first with $77\%$ wins. The right plot shows that LAN bests the other algorithms on 3 maps, namely \texttt{corridor}, \texttt{3s5z\_vs\_3s6z}, \texttt{5m\_vs\_6m}.

\subsection{Credit assignment analysis}

\rav{In the most difficult maps of SMAC the enemy teams have more units and the contribution of all the agents is required to win. The difference of performance between \texttt{3s5z} and \texttt{3s5z\_vs\_3s6z} (same team of agents but one more enemy) is a good example of that. The baiting strategy discovered in \texttt{3s5z\_vs\_3s6z} and \texttt{corridor} showcase the credit assignment of LAN. Indeed, while the agent that serves as bait acts at the beginning of the episode the correct behavior is reinforced even though the rewards for killing the enemies and for defeating the enemy team arrives later.}

\begin{figure}[h!]
    \centering
    \begin{subfigure}[t]{0.3\textwidth}
         \centering
         \includegraphics[width=\textwidth]{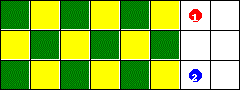}
         \caption{Initial state of the environment.}
         \label{fig:checkers_start}
     \end{subfigure}
     \hspace{2em}
    \begin{subfigure}[t]{0.3\textwidth}
         \centering
         \includegraphics[width=\textwidth]{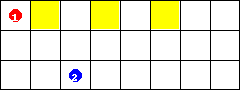}
         \caption{Final state of the environment reached by LAN's policy, with the first agent having eaten all the apples.}
         \label{fig:checkers_end}
     \end{subfigure}
    
    \caption{The Checkers environment. The green boxes are the apples, they yield $+10$ rewards when eaten by the first agent and $+1$ when eaten by the second agent. The yellow boxes are the lemons that yields $-10$ and $-1$ to the first and second agent respectively. The supplementary material contains a gif of the policy.}
    \label{fig:checkers}
\end{figure}

To further emphasize this, we performed an additional experiment on Checkers, an environments of VDN designed to asses credit assignment. In Checkers the red agent gets $+10$ rewards for eating apples (green) and $-10$ rewards for eating lemons (yellow), while the second agent gets $+1$ and $-1$ respectively. The agents receive the sum of both rewards. Each agent receives as observation its location in the map and a $3x3$ window around it. The environment finishes when there are no more apples or after $100$ steps.
Agent 2 needs to eat the lemons (-1 reward) that block the way for agent 1 to eat the apples ($+10$ reward), as shown by the initial state of the environment (Figure \ref{fig:checkers_start}). While the agents get the same team reward, they have distinctive roles as the second agent needs to learn that negative immediate rewards lead to a better team return. LAN converges to the policies described above, with the 3 lemons on the top row left uneaten (Figure \ref{fig:checkers_end}). As this environment was designed to assess the credit assignment problem, this shows that LAN mitigates it. 

\subsection{Moving target problem analysis}
\label{subseq:movingtarget}

IQL serves as a natural ablation of LAN, wherein the shared centralized Value component of our Q-Value proxy is swapped. As discussed in the preceding section, the primary drawback of IQL lies in its susceptibility to the moving target problem, as it disregards the learning of other agents. Consequently, IQL lacks any mitigation strategy against this issue. In scenarios such as \texttt{bane\_vs\_bane}, where coordination is unnecessary or when agents have no mutual influence, IQL can exhibit satisfactory performance. However, the notable superiority of LAN over IQL across all maps demonstrates that LAN effectively addresses the limitations of IQL, including the challenge posed by the moving target problem.

Since the centralized Value of LAN allows to break partial observability we carried out an additional experiment to make sure that the increased performance of LAN was not only due to targets with increased observability. In this experiment, we trained IQL without the fog of war so that all the agents could observe the entire map. While the RNN is no longer needed we kept the same architecture and training procedure of replaying full episodes. This experiment is labelled as "IQL (no fog)" in Figures \ref{fig:maps} and \ref{fig:perf}. In all the maps IQL performs better than IQL without the fog of war. This shows that LAN's performance is not only due to the increased observability of its centralized component and strengthens our claim that our Q-Value proxy mitigates the moving target problem.

In summary, LAN performs on par with the SOTA on the easy and hard maps while dominating the other methods on the super hard maps, even the ones where the other methods did not achieve any wins. LAN outperforms QPLEX by $10\%$ in averaged performance. These results showcase LAN's performance and scalability potential, and its capacity to handle many agents and large observation and action spaces.

\section{Conclusion}
In this paper, we proposed Local Advantage Networks (LAN); a novel value-based MARL algorithm for Dec-POMDPs. 
LAN leverages the CTDE approach by building, for each agent, a proxy of the local Q-value composed of the local advantage and the joint value. LAN trains both networks by applying DQN to a Q-value proxy. The centralized learning allows to condition the joint value on the real state to overcome the partial observability during training. In parallel, it learns the advantages together with the joint value, to synchronize all value functions to the ever changing policies. This results in more accurate DQN targets and mitigates the moving target problem. Conditioning the local advantages solely on the agent's observation-action history, ensures decentralized execution. To ensure scalability, LAN's joint value efficiently summarizes the hidden states produced by the GRUs of the local advantages to represent the joint history. Therefore, the number of parameters of this value function is independent of the number of agents. 

We evaluated LAN on the challenging SMAC benchmark where we performed significantly better or on par compared to state-of-the-art methods, while its architecture is significantly more scalable in the number of agents. In the two most complex maps, LAN was able to learn a complex strategy where one agent would sacrifice itself for the survival of the team, and therefore proving experimentally LAN's ability to mitigate the multi-agent credit assignment problem.
We believe that the lean architecture of LAN for learning decentralized policies in a Dec-POMDP is key to learning efficiently in decentralized partially observable settings.

Most of the recent work in value-based Deep MARL for Dec-POMDP focused on improving the value factorization of QMIX. The need for a different research direction is therefore real, and LAN, by moving away from value factorization, offers an alternative. LAN is not only able to achieve better performance than value factorization but is also more scalable parameter-wise. 

\paragraph{Future work}
In future work, we aim to explore how the history representation of the centralized value can be improved through the use of Attention \citep{Vaswani2017AttentionNeed} or Graph Neural Networks \citep{Kipf2017Semi-supervisedNetworks}. We also aim to investigate  how explicit communication \citep{Oliehoek2008ExploitingDec-POMDPs,Messias2011EfficientPOMDPs,Wang2020R-MADDPGCommunication, Das2019TarMAC:Communication} can be added to LAN to further improve the coordination between the agents and to improve robustness of the learned policies. We also plan to investigate how LAN's architecture might benefit MARL algorithms in settings with continuous action spaces. 

\section*{Acknowledgements}
R. Avalos is supported by the Research Foundation – Flanders (FWO), under grant number 11F5721N.
We thank Florent Delgrange and the anonymous reviewers for their valuable feedback.

\bibliography{references}

\begin{thebibliography}{42}
\providecommand{\natexlab}[1]{#1}
\providecommand{\url}[1]{\texttt{#1}}
\expandafter\ifx\csname urlstyle\endcsname\relax
  \providecommand{\doi}[1]{doi: #1}\else
  \providecommand{\doi}{doi: \begingroup \urlstyle{rm}\Url}\fi

\bibitem[Bu{\c{s}}oniu et~al.(2008)Bu{\c{s}}oniu, Babu{\v{s}}ka, and
  De~Schutter]{Busoniu2008ALearning}
Lucian Bu{\c{s}}oniu, Robert Babu{\v{s}}ka, and Bart De~Schutter.
\newblock {A comprehensive survey of multiagent reinforcement learning}.
\newblock \emph{IEEE Transactions on Systems, Man and Cybernetics Part C:
  Applications and Reviews}, 38\penalty0 (2):\penalty0 156--172, 3 2008.
\newblock \doi{10.1109/TSMCC.2007.913919}.

\bibitem[Cho et~al.(2014)Cho, van Merrienboer, Gulcehre, Bahdanau, Bougares,
  Schwenk, and Bengio]{Cho2014LearningTranslation}
Kyunghyun Cho, Bart van Merrienboer, Caglar Gulcehre, Dzmitry Bahdanau, Fethi
  Bougares, Holger Schwenk, and Yoshua Bengio.
\newblock {Learning Phrase Representations using RNN Encoder-Decoder for
  Statistical Machine Translation}.
\newblock \emph{EMNLP 2014 - 2014 Conference on Empirical Methods in Natural
  Language Processing, Proceedings of the Conference}, pp.\  1724--1734, 6
  2014.
\newblock URL \url{https://arxiv.org/abs/1406.1078v3}.

\bibitem[Das et~al.(2019)Das, Gervet, Romoff, Batra, Parikh, Rabbat, and
  Pineau]{Das2019TarMAC:Communication}
Abhishek Das, Théophile Gervet, Joshua Romoff, Dhruv Batra, Devi Parikh,
  Michael Rabbat, and Joelle Pineau.
\newblock {TarMAC: Targeted multi-agent communication}.
\newblock In \emph{36th International Conference on Machine Learning, ICML
  2019}, volume 2019-June, pp.\  2776--2784, 10 2019.
\newblock ISBN 9781510886988.
\newblock URL \url{http://arxiv.org/abs/1810.11187}.

\bibitem[de~Witt et~al.(2020)de~Witt, Gupta, Makoviichuk, Makoviychuk, Torr,
  Sun, and Whiteson]{deWitt2020IsChallenge}
Christian~Schroeder de~Witt, Tarun Gupta, Denys Makoviichuk, Viktor
  Makoviychuk, Philip H.~S. Torr, Mingfei Sun, and Shimon Whiteson.
\newblock {Is Independent Learning All You Need in the StarCraft Multi-Agent
  Challenge?}
\newblock 11 2020.
\newblock URL \url{http://arxiv.org/abs/2011.09533}.

\bibitem[Foerster et~al.(2017)Foerster, Nardell, Farquhar, Afouras, Torr,
  Kohli, and Whiteson]{Foerster2017StabilisingLearning}
Jakob Foerster, Nantas Nardell, Gregory Farquhar, Trtantafyllos Afouras,
  Philip~H.S. Torr, Pushmeet Kohli, and Shimon Whiteson.
\newblock {Stabilising experience replay for deep multi-agent reinforcement
  learning}.
\newblock In \emph{34th International Conference on Machine Learning, ICML
  2017}, volume~3, pp.\  1879--1888. International Machine Learning Society
  (IMLS), 2 2017.
\newblock ISBN 9781510855144.
\newblock URL \url{http://arxiv.org/abs/1702.08887}.

\bibitem[Foerster et~al.(2018)Foerster, Farquhar, Afouras, Nardelli, and
  Whiteson]{Foerster2018CounterfactualGradients}
Jakob~N. Foerster, Gregory Farquhar, Triantafyllos Afouras, Nantas Nardelli,
  and Shimon Whiteson.
\newblock {Counterfactual multi-agent policy gradients}.
\newblock In \emph{32nd AAAI Conference on Artificial Intelligence, AAAI 2018},
  pp.\  2974--2982, 5 2018.
\newblock ISBN 9781577358008.
\newblock URL \url{http://arxiv.org/abs/1705.08926}.

\bibitem[Hernandez-Leal et~al.(2019)Hernandez-Leal, Kartal, and
  Taylor]{Hernandez-Leal2019ALearning}
Pablo Hernandez-Leal, Bilal Kartal, and Matthew~E. Taylor.
\newblock {A survey and critique of multiagent deep reinforcement learning}.
\newblock \emph{Autonomous Agents and Multi-Agent Systems 2019 33:6},
  33\penalty0 (6):\penalty0 750--797, 10 2019.
\newblock ISSN 1573-7454.
\newblock \doi{10.1007/S10458-019-09421-1}.
\newblock URL
  \url{https://link.springer.com/article/10.1007/s10458-019-09421-1}.

\bibitem[Hochreiter \& Schmidhuber(1997)Hochreiter and
  Schmidhuber]{Hochreiter1997LongMemory}
Sepp Hochreiter and Jürgen Schmidhuber.
\newblock {Long Short-Term Memory}.
\newblock \emph{Neural Computation}, 9\penalty0 (8):\penalty0 1735--1780, 11
  1997.
\newblock ISSN 0899-7667.
\newblock \doi{10.1162/NECO.1997.9.8.1735}.
\newblock URL
  \url{http://direct.mit.edu/neco/article-pdf/9/8/1735/813796/neco.1997.9.8.1735.pdf}.

\bibitem[Huang(2020)]{DBLP:conf/nips/Huang20}
Bojun Huang.
\newblock Steady state analysis of episodic reinforcement learning.
\newblock In Hugo Larochelle, Marc'Aurelio Ranzato, Raia Hadsell,
  Maria{-}Florina Balcan, and Hsuan{-}Tien Lin (eds.), \emph{Advances in Neural
  Information Processing Systems 33: Annual Conference on Neural Information
  Processing Systems 2020, NeurIPS 2020, December 6-12, 2020, virtual}, 2020.
\newblock URL
  \url{https://proceedings.neurips.cc/paper/2020/hash/69bfa2aa2b7b139ff581a806abf0a886-Abstract.html}.

\bibitem[Kipf \& Welling(2017)Kipf and
  Welling]{Kipf2017Semi-supervisedNetworks}
Thomas~N. Kipf and Max Welling.
\newblock {Semi-supervised classification with graph convolutional networks}.
\newblock In \emph{5th International Conference on Learning Representations,
  ICLR 2017 - Conference Track Proceedings}, 9 2017.
\newblock URL \url{http://arxiv.org/abs/1609.02907}.

\bibitem[Klima et~al.(2018)Klima, Bloembergen, Savani, Tuyls, Wittig, Sapera,
  and Izzo]{Klima2018SpaceAnarchy}
Richard Klima, Daan Bloembergen, Rahul Savani, Karl Tuyls, Alexander Wittig,
  Andrei Sapera, and Dario Izzo.
\newblock {Space debris removal: Learning to cooperate and the price of
  anarchy}.
\newblock \emph{Frontiers Robotics AI}, 5\penalty0 (JUN), 2018.
\newblock \doi{10.3389/FROBT.2018.00054/FULL}.

\bibitem[Lowe et~al.(2017)Lowe, Wu, Tamar, Harb, Abbeel, and
  Mordatch]{Lowe2017Multi-agentEnvironments}
Ryan Lowe, Yi~Wu, Aviv Tamar, Jean Harb, Pieter Abbeel, and Igor Mordatch.
\newblock {Multi-agent actor-critic for mixed cooperative-competitive
  environments}.
\newblock In \emph{Advances in Neural Information Processing Systems}, volume
  2017-Decem, pp.\  6380--6391, 6 2017.
\newblock URL \url{http://arxiv.org/abs/1706.02275}.

\bibitem[Lyu et~al.(2021)Lyu, Xiao, Daley, and
  Amato]{Lyu2021ContrastingLearning}
Xueguang Lyu, Yuchen Xiao, Brett Daley, and Christopher Amato.
\newblock {Contrasting Centralized and Decentralized Critics in Multi-Agent
  Reinforcement Learning}.
\newblock \emph{Proc. of the 20th International Conference on AutonomousAgents
  and multi-agent Systems (AAMAS 2021)}, 2 2021.
\newblock URL \url{http://arxiv.org/abs/2102.04402}.

\bibitem[Mahajan et~al.(2019)Mahajan, Rashid, Samvelyan, and
  Whiteson]{Mahajan2019MAVEN:Exploration}
Anuj Mahajan, Tabish Rashid, Mikayel Samvelyan, and Shimon Whiteson.
\newblock {MAVEN: Multi-Agent Variational Exploration}.
\newblock \emph{Advances in Neural Information Processing Systems}, 32, 10
  2019.
\newblock URL \url{https://arxiv.org/abs/1910.07483v2}.

\bibitem[Messias et~al.(2011)Messias, Spaan, and
  Lima]{Messias2011EfficientPOMDPs}
João~V. Messias, Matthijs~T.J. Spaan, and Pedro~U. Lima.
\newblock {Efficient offline communication policies for factored Multiagent
  POMDPs}.
\newblock In \emph{Advances in Neural Information Processing Systems 24: 25th
  Annual Conference on Neural Information Processing Systems 2011, NIPS 2011},
  2011.

\bibitem[Mihaylov et~al.(2010)Mihaylov, Tuyls, and
  Now{\'{e}}]{Mihaylov2010DecentralizedNetworks}
Mihail Mihaylov, Karl Tuyls, and Ann Now{\'{e}}.
\newblock {Decentralized Learning in Wireless Sensor Networks}.
\newblock In Matthew Taylor and Karl Tuyls (eds.), \emph{Adaptive and Learning
  Agents}, pp.\  60--73, Berlin, Heidelberg, 2010. Springer Berlin Heidelberg.
\newblock ISBN 978-3-642-11814-2.

\bibitem[Mnih et~al.(2013)Mnih, Kavukcuoglu, Silver, Graves, Antonoglou,
  Wierstra, and Riedmiller]{Mnih2013PlayingLearning}
Volodymyr Mnih, Koray Kavukcuoglu, David Silver, Alex Graves, Ioannis
  Antonoglou, Daan Wierstra, and Martin Riedmiller.
\newblock {Playing Atari with Deep Reinforcement Learning}.
\newblock 12 2013.
\newblock URL \url{http://arxiv.org/abs/1312.5602}.

\bibitem[Mnih et~al.(2015)Mnih, Kavukcuoglu, Silver, Rusu, Veness, Bellemare,
  Graves, Riedmiller, Fidjeland, Ostrovski, Petersen, Beattie, Sadik,
  Antonoglou, King, Kumaran, Wierstra, Legg, and
  Hassabis]{Mnih2015Human-levelLearning}
Volodymyr Mnih, Koray Kavukcuoglu, David Silver, Andrei~A. Rusu, Joel Veness,
  Marc~G. Bellemare, Alex Graves, Martin Riedmiller, Andreas~K. Fidjeland,
  Georg Ostrovski, Stig Petersen, Charles Beattie, Amir Sadik, Ioannis
  Antonoglou, Helen King, Dharshan Kumaran, Daan Wierstra, Shane Legg, and
  Demis Hassabis.
\newblock {Human-level control through deep reinforcement learning}.
\newblock \emph{Nature}, 518\penalty0 (7540):\penalty0 529--533, 2 2015.
\newblock ISSN 14764687.
\newblock \doi{10.1038/nature14236}.

\bibitem[Now{\'{e}} et~al.(2012)Now{\'{e}}, Vrancx, and
  De~Hauwere]{Nowe2012GameLearning}
Ann Now{\'{e}}, Peter Vrancx, and Yann~Michaël De~Hauwere.
\newblock {Game theory and multi-agent reinforcement learning}.
\newblock In \emph{Adaptation, Learning, and Optimization}. 2012.
\newblock \doi{10.1007/978-3-642-27645-3{\_}14}.

\bibitem[Oliehoek \& Amato(2016)Oliehoek and Amato]{Oliehoek2016APOMDPs}
Frans~A Oliehoek and Christopher Amato.
\newblock \emph{{A Concise Introduction to Decentralized POMDPs}}.
\newblock 2016.
\newblock ISBN 978-3-319-28927-4.
\newblock URL \url{http://www.springer.com/us/book/
  http://link.springer.com/10.1007/978-3-319-28929-8
  http://www.springer.com/us/book/%0Ahttp://link.springer.com/10.1007/978-3-319-28929-8}.

\bibitem[Oliehoek et~al.(2008{\natexlab{a}})Oliehoek, Spaan, and
  Vlassis]{Oliehoek2008OptimalPOMDPs}
Frans~A. Oliehoek, Matthijs~T.J. Spaan, and Nikos Vlassis.
\newblock {Optimal and approximate Q-value functions for decentralized POMDPs}.
\newblock \emph{Journal of Artificial Intelligence Research}, 32:\penalty0
  289--353, 10 2008{\natexlab{a}}.
\newblock ISSN 10769757.
\newblock \doi{10.1613/jair.2447}.
\newblock URL \url{http://dx.doi.org/10.1613/jair.2447}.

\bibitem[Oliehoek et~al.(2008{\natexlab{b}})Oliehoek, Spaan, Whiteson, and
  Vlassis]{Oliehoek2008ExploitingDec-POMDPs}
Frans~A. Oliehoek, Matthijs~T.J. Spaan, Shimon Whiteson, and Nikos Vlassis.
\newblock {Exploiting locality of interaction in factored Dec-POMDPs}.
\newblock In \emph{Proceedings of the International Joint Conference on
  Autonomous Agents and Multiagent Systems, AAMAS}, volume~1,
  2008{\natexlab{b}}.

\bibitem[Oliehoek et~al.(2012)Oliehoek, Witwicki, and
  Kaelbling]{Oliehoek2012Influence-basedSystems}
Frans~A. Oliehoek, Stefan~J. Witwicki, and Leslie~P. Kaelbling.
\newblock {Influence-based abstraction for multiagent systems}.
\newblock In \emph{Proceedings of the National Conference on Artificial
  Intelligence}, 2012.
\newblock ISBN 9781577355687.

\bibitem[Rashid et~al.(2018)Rashid, Samvelyan, de~Witt, Farquhar, Foerster, and
  Whiteson]{Rashid2018QMIX:Learning}
Tabish Rashid, Mikayel Samvelyan, Christian~Schroeder de~Witt, Gregory
  Farquhar, Jakob Foerster, and Shimon Whiteson.
\newblock {QMIX: Monotonic Value Function Factorisation for Deep Multi-Agent
  Reinforcement Learning}.
\newblock \emph{Proceedings of the35thInternational Conference on
  MachineLearning, Stockholm, Sweden, PMLR 80, 2018}, 3 2018.
\newblock URL \url{https://arxiv.org/abs/1803.11485v2}.

\bibitem[Samvelyan et~al.(2019)Samvelyan, Rashid, de~Witt, Farquhar, Nardelli,
  Rudner, Hung, Torr, Foerster, and Whiteson]{Samvelyan2019TheChallenge}
Mikayel Samvelyan, Tabish Rashid, Christian~Schroeder de~Witt, Gregory
  Farquhar, Nantas Nardelli, Tim G.~J. Rudner, Chia-Man Hung, Philip H.~S.
  Torr, Jakob Foerster, and Shimon Whiteson.
\newblock {The StarCraft Multi-Agent Challenge}.
\newblock \emph{Proceedings of the International Joint Conference on Autonomous
  Agents and Multiagent Systems, AAMAS}, 4:\penalty0 2186--2188, 2 2019.
\newblock URL \url{https://arxiv.org/abs/1902.04043v5}.

\bibitem[Shoham et~al.(2007)Shoham, Powers, and Grenager]{Shoham2007IfQuestion}
Yoav Shoham, Rob Powers, and Trond Grenager.
\newblock {If multi-agent learning is the answer, what is the question?}
\newblock \emph{Artificial Intelligence}, 2007.
\newblock ISSN 00043702.
\newblock \doi{10.1016/j.artint.2006.02.006}.

\bibitem[Silver et~al.(2016)Silver, Huang, Maddison, Guez, Sifre, Driessche,
  Schrittwieser, Antonoglou, Panneershelvam, Lanctot, Dieleman, Grewe, Nham,
  Kalchbrenner, Sutskever, Lillicrap, Leach, Kavukcuoglu, Graepel, and
  Hassabis]{Silver2016MasteringSearch}
David Silver, Aja Huang, Christopher~J. Maddison, Arthur Guez, Laurent Sifre,
  George van~den Driessche, Julian Schrittwieser, Ioannis Antonoglou, Veda
  Panneershelvam, Marc Lanctot, Sander Dieleman, Dominik Grewe, John Nham, Nal
  Kalchbrenner, Ilya Sutskever, Timothy Lillicrap, Madeleine Leach, Koray
  Kavukcuoglu, Thore Graepel, and Demis Hassabis.
\newblock {Mastering the game of Go with deep neural networks and tree search}.
\newblock 2016.

\bibitem[Sunehag et~al.(2018)Sunehag, Lever, Gruslys, Czarnecki, Zambaldi,
  Jaderberg, Lanctot, Sonnerat, Leibo, Tuyls, and
  Graepel]{Sunehag2018Value-decompositionReward}
Peter Sunehag, Guy Lever, Audrunas Gruslys, Wojciech~Marian Czarnecki, Vinicius
  Zambaldi, Max Jaderberg, Marc Lanctot, Nicolas Sonnerat, Joel~Z. Leibo, Karl
  Tuyls, and Thore Graepel.
\newblock {Value-decomposition networks for cooperative multi-agent learning
  based on team reward}.
\newblock In \emph{Proceedings of the International Joint Conference on
  Autonomous Agents and Multiagent Systems, AAMAS}, volume~3, pp.\  2085--2087,
  6 2018.
\newblock ISBN 9781510868083.
\newblock URL \url{http://arxiv.org/abs/1706.05296}.

\bibitem[Sutton \& Barto(1998)Sutton and
  Barto]{Sutton1998ReinforcementIntroduction}
R.S. Sutton and A.G. Barto.
\newblock {Reinforcement Learning: An Introduction}.
\newblock \emph{IEEE Transactions on Neural Networks}, 1998.
\newblock ISSN 1045-9227.
\newblock \doi{10.1109/tnn.1998.712192}.

\bibitem[Tampuu et~al.(2015)Tampuu, Matiisen, Kodelja, Kuzovkin, Korjus, Aru,
  Aru, and Vicente]{Tampuu2015MultiagentLearning}
Ardi Tampuu, Tambet Matiisen, Dorian Kodelja, Ilya Kuzovkin, Kristjan Korjus,
  Juhan Aru, Jaan Aru, and Raul Vicente.
\newblock {Multiagent Cooperation and Competition with Deep Reinforcement
  Learning}.
\newblock \emph{PLoS ONE}, 12\penalty0 (4), 11 2015.
\newblock URL \url{http://arxiv.org/abs/1511.08779}.

\bibitem[Tan(1993)]{Tan1993Multi-AgentAgents}
Ming Tan.
\newblock {Multi-Agent Reinforcement Learning: Independent vs. Cooperative
  Agents}.
\newblock In \emph{Machine Learning Proceedings 1993}. 1993.
\newblock \doi{10.1016/b978-1-55860-307-3.50049-6}.

\bibitem[Tuyls \& Weiss(2012)Tuyls and Weiss]{Tuyls2012MultiagentProspects}
Karl Tuyls and Gerhard Weiss.
\newblock {Multiagent learning: Basics, challenges, and prospects}.
\newblock In \emph{AI Magazine}, 2012.
\newblock \doi{10.1609/aimag.v33i3.2426}.

\bibitem[van Hasselt et~al.(2015)van Hasselt, Guez, and
  Silver]{vanHasselt2015DeepQ-learning}
Hado van Hasselt, Arthur Guez, and David Silver.
\newblock {Deep Reinforcement Learning with Double Q-learning}.
\newblock \emph{30th AAAI Conference on Artificial Intelligence, AAAI 2016},
  pp.\  2094--2100, 9 2015.
\newblock URL \url{https://arxiv.org/abs/1509.06461v3}.

\bibitem[Vaswani et~al.(2017)Vaswani, Shazeer, Parmar, Uszkoreit, Jones, Gomez,
  Kaiser, and Polosukhin]{Vaswani2017AttentionNeed}
Ashish Vaswani, Noam Shazeer, Niki Parmar, Jakob Uszkoreit, Llion Jones,
  Aidan~N. Gomez, Lukasz Kaiser, and Illia Polosukhin.
\newblock {Attention is all you need}.
\newblock In \emph{Advances in Neural Information Processing Systems}, volume
  2017-Decem, pp.\  5999--6009, 2017.

\bibitem[Wang et~al.(2021)Wang, Ren, Liu, Yu, and
  Zhang]{Wang2021QPLEX:Q-Learning}
Jianhao Wang, Zhizhou Ren, Terry Liu, Yang Yu, and Chongjie Zhang.
\newblock {QPLEX: Duplex Dueling Multi-Agent Q-Learning}.
\newblock \emph{International Conference on Learning Representations}, 8 2021.
\newblock URL \url{https://arxiv.org/abs/2008.01062
  http://arxiv.org/abs/2008.01062 https://openreview.net/forum?id=Rcmk0xxIQV}.

\bibitem[Wang et~al.(2020)Wang, Everett, and
  How]{Wang2020R-MADDPGCommunication}
Rose~E. Wang, Michael Everett, and Jonathan~P. How.
\newblock {R-MADDPG for Partially Observable Environments and Limited
  Communication}.
\newblock 2 2020.
\newblock ISSN 2331-8422.
\newblock URL \url{http://arxiv.org/abs/2002.06684}.

\bibitem[Wang \& Dong(2020)Wang and Dong]{Wang2020ROMA:Roles}
Tonghan Wang and Heng Dong.
\newblock {Roma: Multi-Agent reinforcement learning with emergent roles}.
\newblock In \emph{37th International Conference on Machine Learning, ICML
  2020}, volume PartF16814, pp.\  9818--9828, 3 2020.
\newblock ISBN 9781713821120.
\newblock URL \url{http://arxiv.org/abs/2003.08039}.

\bibitem[Wang et~al.(2016)Wang, Schaul, Hessel, Van~Hasselt, Lanctot, and
  De~Frcitas]{Wang2016DuelingLearning}
Ziyu Wang, Tom Schaul, Matteo Hessel, Hado Van~Hasselt, Marc Lanctot, and Nando
  De~Frcitas.
\newblock {Dueling Network Architectures for Deep Reinforcement Learning}.
\newblock In \emph{33rd International Conference on Machine Learning, ICML
  2016}, volume~4, pp.\  2939--2947, 11 2016.
\newblock ISBN 9781510829008.
\newblock URL \url{https://arxiv.org/abs/1511.06581}.

\bibitem[Xu et~al.(2020)Xu, Gholami, Carthy, Dilkina, Plumptre, Tambe, Singh,
  Nsubuga, Mabonga, Driciru, Wanyama, Rwetsiba, Okello, and
  Enyel]{Xu2020StayVersion}
Lily Xu, Shahrzad Gholami, Sara~Mc Carthy, Bistra Dilkina, Andrew Plumptre,
  Milind Tambe, Rohit Singh, Mustapha Nsubuga, Joshua Mabonga, Margaret
  Driciru, Fred Wanyama, Aggrey Rwetsiba, Tom Okello, and Eric Enyel.
\newblock {Stay ahead of poachers: Illegal wildlife poaching prediction and
  patrol planning under uncertainty with field test evaluations (Short
  Version)}.
\newblock \emph{Proceedings - International Conference on Data Engineering},
  2020-April:\penalty0 1898--1901, 4 2020.
\newblock \doi{10.1109/ICDE48307.2020.00198}.

\bibitem[Yang et~al.(2020)Yang, Hao, Liao, Shao, Chen, Liu, and
  Tang]{Yang2020Qatten:Learning}
Yaodong Yang, Jianye Hao, Ben Liao, Kun Shao, Guangyong Chen, Wulong Liu, and
  Hongyao Tang.
\newblock {Qatten: A General Framework for Cooperative Multiagent Reinforcement
  Learning}.
\newblock 2 2020.
\newblock URL \url{https://arxiv.org/abs/2002.03939v2
  http://arxiv.org/abs/2002.03939}.

\bibitem[Yu et~al.(2021)Yu, Velu, Vinitsky, Wang, Bayen, and
  Wu]{Yu2021TheGames}
Chao Yu, Akash Velu, Eugene Vinitsky, Yu~Wang, Alexandre Bayen, and Yi~Wu.
\newblock {The Surprising Effectiveness of PPO in Cooperative, Multi-Agent
  Games}.
\newblock 3 2021.
\newblock URL \url{http://arxiv.org/abs/2103.01955}.

\bibitem[Zahavy et~al.(2018)Zahavy, Haroush, Merlis, Mankowitz, and
  Mannor]{Zahavy2018LearnLearning}
Tom Zahavy, Matan Haroush, Nadav Merlis, Daniel~J. Mankowitz, and Shie Mannor.
\newblock {Learn what not to learn: Action elimination with deep reinforcement
  learning}.
\newblock In \emph{Advances in Neural Information Processing Systems}, volume
  2018-Decem, pp.\  3562--3573, 9 2018.
\newblock URL \url{http://arxiv.org/abs/1809.02121}.

\end{thebibliography}
\bibliographystyle{tmlr}
\newpage
\appendix
\section{StarCraft Multi-Agent Challenge}
\label{app:smac}

The complete information about the SMAC benchmark can be found in the introductory paper \citep{Samvelyan2019TheChallenge}. Table \ref{tab:my-table} lists the 14 different maps of the challenge with the number of agents in each team and the number of parameters of the centralized part of LAN, QPLEX and QMIX. Table \ref{tab:param_table_full} lists the number of parameters of the centralized component of LAN, QMIX and QPLEX for the 14 maps.

\begin{table}[h]
\caption{The different maps of SMAC.}
\begin{center}
\begin{tabular}{ccc}
\hline
\textbf{Map Name}  & \textbf{Ally Units}                 & \textbf{Enemy Units}                \\ \hline
2s3z               & 2 Stalkers \& 3 Zealots             & 2 Stalkers \& 3 Zealots             \\ 
3s5z               & 3 Stalkers \& 5 Zealots             & 3 Stalkers \& 5 Zealots             \\ 
1c3s5z             & 1 Colossus, 3 Stalkers \& 5 Zealots & 1 Colossus, 3 Stalkers \& 5 Zealots \\ \hline
5m\_vs\_6m         & 5 Marines                           & 6 Marines                           \\ 
10m\_vs\_11m       & 10 Marines                          & 11 Marines                          \\ 
27m\_vs\_30m       & 27 Marines                          & 30 Marines                          \\ 
3s5z\_vs\_3s6z     & 3 Stalkers \& 5 Zealots             & 3 Stalkers \& 6 Zealots             \\ 
MMM2               & 1 Medivac, 2 Marauders \& 7 Marines & 1 Medivac, 3 Marauders \& 8 Marines \\ \hline
2s\_vs\_1sc        & 2 Stalkers                          & 1 Spine Crawler                     \\ 
3s\_vs\_5z         & 3 Stalkers                          & 5 Zealots                           \\ 
6h\_vs\_8z         & 6 Hydralisks                        & 8 Zealots                           \\ 
bane\_vs\_bane     & 20 Zerglings \& 4 Banelings         & 20 Zerglings \& 4 Banelings         \\ 
2c\_vs\_64zg       & 2 Colossi                           & 64 Zerglings                        \\ 
corridor           & 6 Zealots                           & 24 Zerglings                        \\ \hline
\end{tabular}
\end{center}

\label{tab:my-table}
\end{table}

\begin{table}[h]
\caption{Number of parameters (x1000) of the value function in LAN vs.\ the mixing network in QPLEX/QMIX.\\}

    \centering
    \begin{tabular}{crrr}
        \toprule
         &  \textbf{LAN} &  \textbf{QPLEX} &  \textbf{QMIX} \\
        \midrule
        \textbf{2s3z}         &   62 &     50 &    36 \\
        \textbf{3s5z}         &   74 &     90 &    60 \\
        \textbf{1c3s5z}       &   83 &    113 &    73 \\\hline
        \textbf{5m\_vs\_6m}     &   56 &     43 &    32 \\
        \textbf{10m\_vs\_11m}   &   68 &    106 &    70 \\
        \textbf{27m\_vs\_30m}   &  111 &    709 &   283 \\
        \textbf{3s5z\_vs\_3s6z} &   76 &     95 &    63 \\
        \textbf{MMM2}         &   86 &    136 &    85 \\\hline
        \textbf{2s\_vs\_1sc}    &   46 &     18 &    12 \\
        \textbf{3s\_vs\_5z}     &   54 &     31 &    22 \\
        \textbf{6h\_vs\_8z}     &   61 &     59 &    42 \\
        \textbf{bane\_vs\_bane} &  125 &    555 &   241 \\
        \textbf{2c\_vs\_64zg}   &  119 &    116 &    72 \\
        \textbf{corridor}     &   79 &    109 &    69 \\
        \bottomrule
        \end{tabular}
    
    \label{tab:param_table_full}
\end{table}

\section{Implementation details}
\label{app:implem}

We use neural networks with ReLu activation functions, to approximate the local advantage and the centralized value. 
To increase the learning speed and reduce the number of parameters we share the neural network weights of the local advantages between all the agents. The input of the advantage network conditions on the agent ID so that the policy can differ per agent. 
The advantage network is composed of a 2 hidden layers, a $64$ units feed forward network and a $64$ units GRU, which is consistent with the architecture used in the SOTA algorithms to represent the decentralized utilities \citep{Rashid2018QMIX:Learning, Wang2021QPLEX:Q-Learning}. %

The centralized value network (Figure \ref{fig:architecture}, left) first computes an embedding of $\Tilde{h}_a$ for each agent, $\hat{h}_a$, using a feed forward network of $128$ units. The agents' embeddings are then merged together by summing them resulting in a joint history embedding of fixed size. This joint history embedding is then concatenated with the real state provided by the environment to create a state-history embedding. Finally, this state-history embedding goes through an feed forward network of two hidden layers of $128$ units to compute the value.

We train LAN for $2$ million timesteps using a replay buffer of $5k$ episodes. During training we use an $\varepsilon$-greedy exploration strategy over the local advantages, with $\varepsilon$ decaying from $1$ to $0.05$ over the first $50k$ timesteps. After every episode we optimize both networks twice using Adam with a learning rate of $5e^{-4}$ and without TD($\lambda$). 
For each update we sample a batch of $32$ episodes from the replay buffer. The DQN target are computed with a target network that is updated every $200$ gradient updates. We clip to $10$ the norm of the gradient.

We note that LAN does not require parameter sharing, and that each type of agent could have its own model. In that case, every agent type also needs its own embedding network to compute $\Tilde{h}_a$.

\section{Remaining maps of SMAC}
\label{app:remaining}

\begin{figure*}[h]
    \centering
    \includegraphics[width=\textwidth]{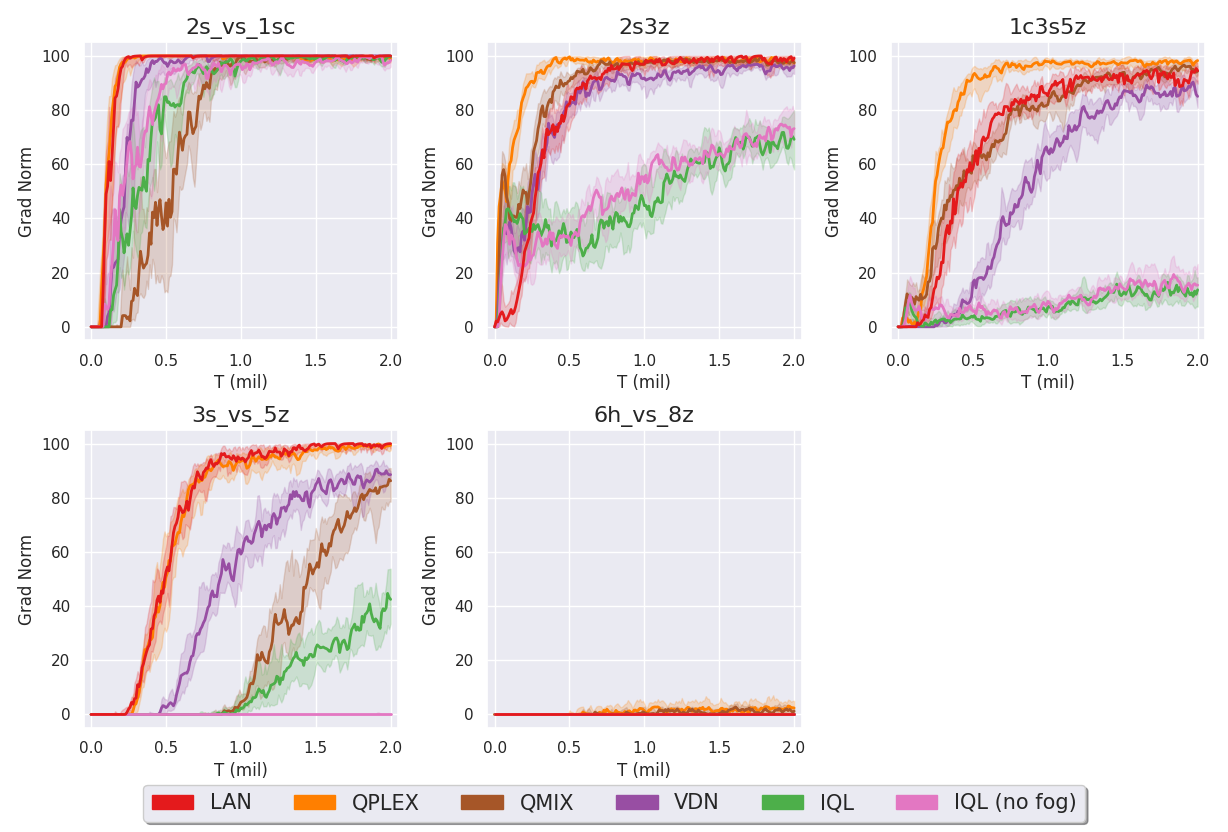}
    \caption{Median battle won rate during learning on the last 5 SMAC maps. %
    }
    \label{fig:additional_plots}
\end{figure*}

Figure \ref{fig:additional_plots} includes the 5 SMAC maps that are not included in the main paper. The first map, \texttt{2s\_vs\_1sc}, is an easy map and LAN learns the perfect strategy as the other algorithms do. In the second and third maps, \texttt{2s3z} and \texttt{1c3s5z}, all the algorithms but IQL learn near-optimal policies. In \texttt{3s\_vs\_5z}, LAN and QPLEX learn the optimal policy followed closely by QMIX and VDN that both reach around $85\%$. 
Finally, in the last map \texttt{6h\_vs\_8z} no algorithm is able to score any wins. We note that the difference in performance between IQL and IQL (no fog) is consistent with the other maps: removing the fog of war does not increase performance. 
\section{Discussion regarding the advantage}
\label{app:adv}

\begin{figure*}[h]
    \centering
    \includegraphics[width=\textwidth]{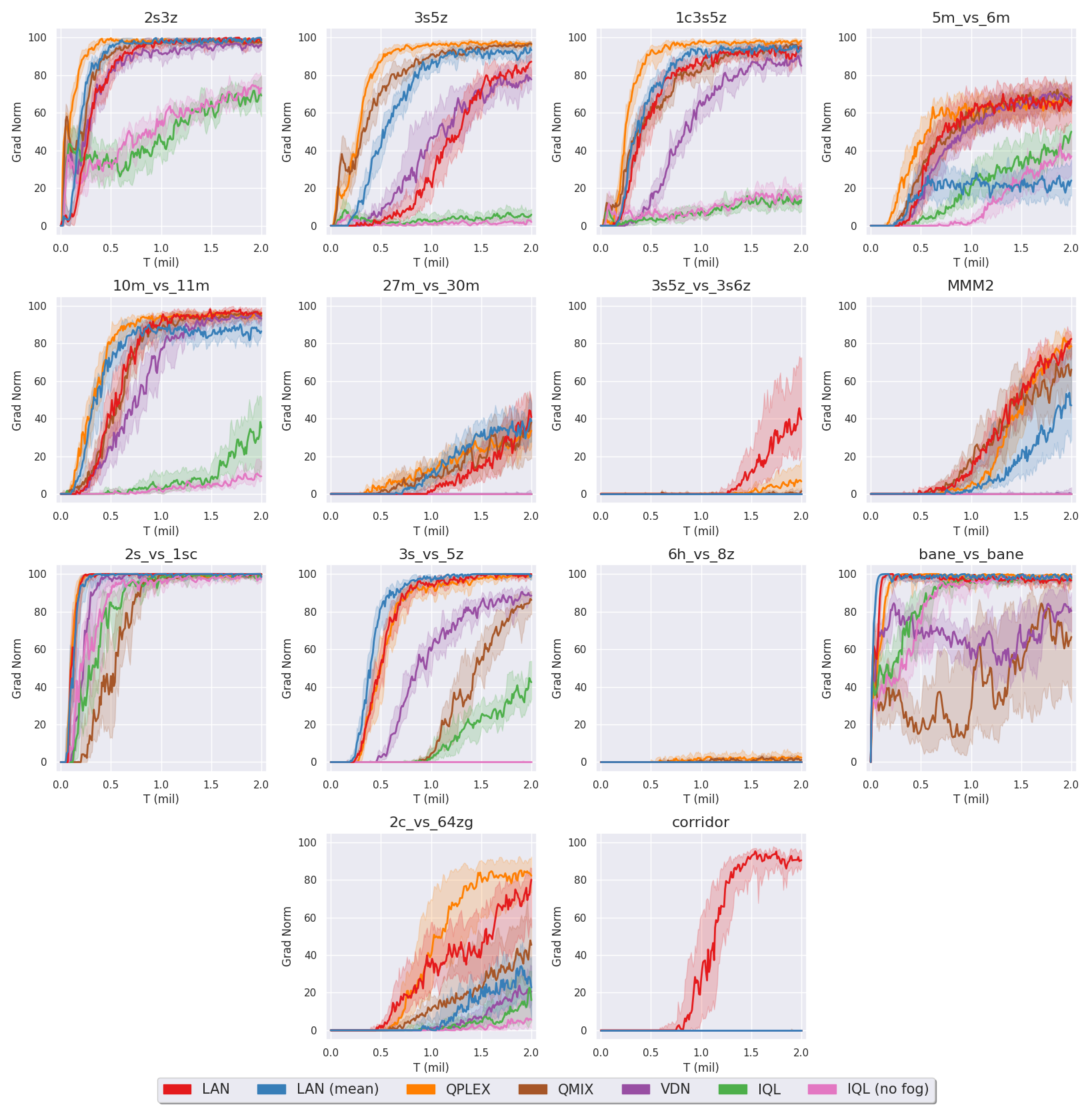}
    \caption{Median battle won rate during learning on the all the SMAC maps.}
    \label{fig:smac_mean}
\end{figure*}

Figure \ref{fig:smac_mean} shows the performance on all the SMAC maps with a variation of LAN called LAN mean, which applies the equation \ref{eq:adv_mean} . While in two maps \texttt{3s5z}, \texttt{27m\_vs\_30m} the mean version of LAN improves over the classical version, it degrades the performance in others other maps such as \texttt{5m\_vs\_6m}, \texttt{2c\_vs\_64zg}, and \texttt{MMM2}, and prevents the learning in \texttt{corridor} and \texttt{3s5z\_vs\_3s6z}. This empirically shows that while in the single agent case the equation \ref{eq:adv_mean} stabilizes the learning it might not be the case when multiple agents are involved.

\section{Algorithm}
\label{sec:alg}
\begin{algorithm}[H]
    \SetKwInOut{Input}{Input}
    \SetKwInOut{Output}{Output}
    \SetKwFunction{Initialize}{Initialize}
    \SetKwFunction{SelectAction}{SelectAction}
    \SetKwFunction{ExecuteAction}{ExecuteJointAction}
    \SetKwFunction{StoreTransition}{StoreTransition}
    \SetKwFunction{UpdateCurrentState}{UpdateCurrentState}
    \SetKwFunction{UpdateCurrentObs}{UpdateCurrentJointObs}
    \SetKwFunction{UpdateTotalReward}{UpdateTotalReward}
    \SetKwFunction{UpdateV}{UpdateValue}
    \SetKwFunction{UpdateA}{UpdateAdvantages}
    \SetKwFunction{UpdateVandA}{UpdateValueAndLocalAdvantages}
    \SetKwFunction{ComputeEpisodeTarget}{ComputeEpisodeTarget}
    \SetKwFunction{UpdateTargetNetwork}{UpdateTargetNetwork}
    \SetKwFunction{PrintEpisodeInfo}{PrintEpisodeInfo}
    \SetKwFunction{ResetEnvironment}{ResetEnvironment}
    \SetKwFunction{ResetHiddenStates}{ResetHiddenStates}
    \SetKwFunction{ResetLastAction}{ResetLastAction}
    \SetKwFunction{UpdateHiddenState}{UpdateHiddenState}
    \SetKwFunction{UpdateHiddenStateAndSelectAction}{UpdateHiddenStateAndSelectAction}
    \SetKwFunction{UpdateExploration}{UpdateExploration}

    \Input{Agent set $N$}
    \Input{Replay memory capacity $MC$}
    \Input{Frequency of target update $C$}
    \Input{Exploration rate $\epsilon$}
    \BlankLine
    \Initialize{replay memory $D$ with capacity $MC$}\;
    \Initialize{centralized value function $V$ with random weights}\;
    \Initialize{for each agent $a \in N$ local advantage function $A^a$, containing $RNN_a$, with random weights}\;
    \Initialize{target value function $V_t$ with weights of $V$, and for each agent $a$ target local advantage $A^a_t$ with weights $A^a$} \;
    \Initialize{epsilon decay $\epsilon_{decay}$ and minimum epsilon $\epsilon_{\min}$}\;
    \BlankLine
    \For{$episode$}{
        \tcp{Interaction with environment}
        \Initialize{empty episode memory $E$}\\
        \ResetEnvironment{$s, \bm{o} \leftarrow env$} \tcp{get state and joint observation}
        \ResetHiddenStates{$\forall a \in N, \tau_{a} = 0$}\;
        \ResetLastAction{$\forall a \in N, u_{a} = 0$}
        
        \While{episode is not finished}{
            
            \For{agent $a \in N$}{
                \UpdateHiddenState{$\tau_a \leftarrow RNN_a(\tau_a, u_a, o_a)$}\;
                \SelectAction{$u_a \leftarrow \pi_{\epsilon}(A_a(\tau_a))$}\;
            }
            \ExecuteAction{$\bm{u}$} \;
            Observe next state $s'$, next joint observation $\bm{o}$, reward $r$ \;
            \StoreTransition{$s$, $\bm{o}$, $\bm{u}$, $r$, $s'$ $\bm{o}'$, in episode memory $E$} \;
            \UpdateCurrentState{$s \leftarrow s'$} \;
            \UpdateCurrentObs{$\bm{o} \leftarrow \bm{o'}$} \;
        }
        
        Store episode memory $E$ in replay memory $D$ \;
        \BlankLine
        \tcp{Perform learning step}

        Sample random batch $B$ of episodes from $D$ \;
        
        \For{each episode $e$ in the batch $B$}{
            \For{each timestep $t=1$ to last step of the episode $T(e)$}{
                Unroll RNN of current and target networks\;
                For each agent $a$ compute current $\Tilde{Q}$ estimate using Equation \ref{eq:q_tilde}\;
                \tcp{$\Tilde{Q}^{\bm{\pi}}_a(s, \bm{\tau}, u_a) = V^{\bm{\pi}}(s, \bm{\tau}) + A^{\pi_a}(\tau_a, u_a)$}
                For each agent $a$ compute TD target with target networks using Equation \ref{eq:target}\;
                \tcp{$y_a = r + \gamma [V^{\bm{\pi}}_t(s', \bm{\tau}') + A^{\pi_a}_t(\tau_a', \arg\max_{u_a'} A^{\pi_a}(\tau_a', u_a'))]$}
                For each agent $a$ compute $TD_{a, e, t}$ the temporal difference error\;
            }
            
        }
        \UpdateVandA{using gradient descent on the mean square temporal difference error}\;
        \BlankLine
        \tcp{Update target network and exploration}
        \UpdateTargetNetwork{$\forall a \in N, A^t_a \leftarrow A; V^t \leftarrow V$} (every $C$ steps) \;
        \UpdateExploration{$\epsilon \leftarrow \max(\epsilon \times \epsilon_{decay}; \epsilon_{\min})$} \;
    }
    \caption{Local Advantage Networks (LAN)}
\end{algorithm}
\newpage
\section{Proof}
\label{sec:proof}
\textbf{Episodic process.} A POMDP $\pomdp$ is episodic if it includes a special \emph{reset state} that is fully observable by the agent, and that under any policy the environment is almost surely eventually reset. Furthermore, when the environment is reset it transition to the initial state. 

For this proof we consider an agent $a$ with policy $\policy_a$ and the induced POMDP $G_a$ obtained by fixing the policy of the other agents $\jointpolicy_{-a}$ (defined in Section \ref{sec:method}). 

Without any loss of generality, we augment $G_a$ with an observable reset state so that $G_a$ is episodic. 
This ensures the ergodicity of $G_a$, as every epsiodic process is ergodic or can be made ergodic without loss of generality \cite{DBLP:conf/nips/Huang20}, and consequently the existence of a stationary distribution $p_{\jointpolicy}\fun{\tilde{\state}, \history_a} = p_{\jointpolicy}\fun{\state, \jointhistory}$.\\\

As LAN learns greedy policies we consider only deterministic policies.

\subsection{Warm-up}

By decomposing the next joint history $\jointhistory'$ as a tuple containing the new joint observation $\jointobservation'$, the joint action $\jointaction$ and the joint history $\jointhistory$ we obtain the following equality:
\begin{align}
    p\fun{\jointhistory'=\tuple{\jointobservation', \jointaction, \tilde{\jointhistory}} \mid \state', \jointpolicy\fun{\jointhistory}, \jointhistory} &= \diracimpulse{\jointhistory}{\tilde{\jointhistory}} \diracimpulse{\jointpolicy\fun{\jointhistory}}{\jointaction}
    \observationfn\fun{\jointobservation' \mid \state', \jointpolicy\fun{\jointhistory}}
\end{align}
Where $\diracimpulse{x}{y}$ is the Kronecker delta symbol. It is equal to $1$ if $x = y$ and $0$ otherwise.

We can obtain a similar result for the next local history $\history_a'$
\begin{align}
    p\fun{\history_a' = \tuple{\observation_a', \action_a, \tilde{\history_a}} \mid \state', \jointpolicy\fun{\tuple{\jointhistory_{-a}, \history_{a}}}, \tuple{\jointhistory_{-a}, \history_{a}}}  &= \diracimpulse{\history_a}{\tilde{\history}_a} \diracimpulse{\policy_a\fun{\history_a}}{\action_a}\observationfn_a\fun{\observation_a' \mid \state', \policy_a\fun{\history_a}}
\end{align}

For any local history $\history_a$ of agent $a$ that is realisable under the policy $\policy_a$ we can define the following conditional probability:
\begin{align}
    p\fun{\state, \jointhistory_{-a} \mid \history_a} &= \frac{p_{\jointpolicy}\fun{\state, \tuple{\jointhistory_{-a}, \history_a}}}{p\fun{\history_a}} = \frac{p_{\jointpolicy}\fun{\state, \tuple{\jointhistory_{-a}, \history_a}}}{
    \expectedsymbol{\state', \tuple{\jointhistory_{-a}', \history_a'} \sim p_{\jointhistory}} \diracimpulse{\history_a}{\history_a'}}
\end{align}

For any realisable history $\history_a$ of agent $a$ that is realisable under the policy $\policy_a$, and next history $\history_a'$ we have:
\begin{align}
    p\fun{\history_a' \mid \history_a} &= \expectedsymbol{\state, \jointhistory_{-a} \sim p\fun{ \cdot \mid \history_a}} \expectedsymbol{\state' \sim \probtransitions\fun{\cdot \mid \state, \jointpolicy\fun{\tuple{\jointhistory_{-a}, \history_{a}}}}} p\fun{\history_a' \mid \state', \tuple{\jointhistory_{-a}, \history_a}, \jointpolicy\fun{\tuple{\jointhistory_{-a}, \history_a}}}
    \label{eq:nextlocalhistorybasedlocalhistory}
\end{align}

\textit{Proof:}
\begin{align*}
    p\fun{\history_a' \mid \history_a} 
    &= \int_{\state} \int_{\state'} \int_{\jointhistory_{-a}} p\fun{\history_a' \mid \state', \tuple{\jointhistory_{-a}, \history_a}, \jointpolicy\fun{\tuple{\jointhistory_{-a}, \history_a}}, \state} p\fun{\state', \jointhistory_{-a}, \jointpolicy\fun{\tuple{\jointhistory_{-a}, \history_a}}, \state \mid \history_a} \text{d}\jointhistory_{-a}\text{d}\state'\text{d}\state \tag{law of total probability} \\
    &= \int_{\state} \int_{\state'} \int_{\jointhistory_{-a}} p\fun{\history_a' \mid \state', \tuple{\jointhistory_{-a}, \history_a}, \jointpolicy\fun{\tuple{\jointhistory_{-a}, \history_a}}, \state} p\fun{\state, \jointhistory_{-a} \mid \history_a} \probtransitions\fun{\state' \mid \state, \tuple{\jointhistory_{-a}, \history_a}, \jointpolicy\fun{\tuple{\jointhistory_{-a}, \history_a}}} 
    \text{d}\jointhistory_{-a}\text{d}\state'\text{d}\state \tag{chain rule}\\
    &= \int_{\state}\int_{\jointhistory_{-a}}  p\fun{\state, \jointhistory_{-a} \mid \history_a} \int_{\state'} \probtransitions\fun{\state' \mid \state, \tuple{\jointhistory_{-a}, \history_a}, \jointpolicy\fun{\tuple{\jointhistory_{-a}, \history_a}}} p\fun{\history_a' \mid \state', \tuple{\jointhistory_{-a}, \history_a}, \jointpolicy\fun{\tuple{\jointhistory_{-a}, \history_a}}, \state} 
    \text{d}\state' \text{d}\state \text{d}\jointhistory_{-a} \tag{linearity}\\
    &= \int_{\state}\int_{\jointhistory_{-a}}  p\fun{\state, \jointhistory_{-a} \mid \history_a} \expectedsymbol{\state' \sim \probtransitions\fun{\cdot \mid \state, \jointpolicy\fun{\tuple{\jointhistory_{-a}, \history_{a}}}}} p\fun{\history_a' \mid \state', \tuple{\jointhistory_{-a}, \history_a}, \jointpolicy\fun{\tuple{\jointhistory_{-a}, \history_a}}, \state} 
     \text{d}\state \text{d}\jointhistory_{-a} \tag{definition of expectation} \\
    &= \expectedsymbol{\state, \jointhistory_{-a} \sim p\fun{ \cdot \mid \history_a}} \expectedsymbol{\state' \sim \probtransitions\fun{\cdot \mid \state, \jointpolicy\fun{\tuple{\jointhistory_{-a}, \history_{a}}}}} p\fun{\history_a' \mid \state', \tuple{\jointhistory_{-a}, \history_a}, \jointpolicy\fun{\tuple{\jointhistory_{-a}, \history_a}}, \state}  \tag{definition of expectation}\\
    &= \expectedsymbol{\state, \jointhistory_{-a} \sim p\fun{ \cdot \mid \history_a}} \expectedsymbol{\state' \sim \probtransitions\fun{\cdot \mid \state, \jointpolicy\fun{\tuple{\jointhistory_{-a}, \history_{a}}}}} p\fun{\history_a' \mid \state', \tuple{\jointhistory_{-a}, \history_a}, \jointpolicy\fun{\tuple{\jointhistory_{-a}, \history_a}}} \tag{conditional independence of $\history_a'$ and $\state$ given $\state'$} \\
\end{align*}

For any realisable history $\history_a$, that is realisable under the policy $\policy_a$, and any next state $\state'$ and next joint history $\jointhistory'$ we have
\begin{align}
    p\fun{\state', \jointhistory' \mid \history_a} 
    &= \expectedsymbol{\state, \jointhistory_{-a} \sim p\fun{\cdot \mid \history_a}} \probtransitions\fun{\state' \mid \state, \jointpolicy\fun{\tuple{\jointhistory_{-a}, \history_a}}} p\fun{\jointhistory' \mid \state', \tuple{\jointhistory_{-a}, \history_a}, \jointpolicy\fun{\tuple{\jointhistory_{-a}, \history_a}}}
    \label{eq:nextstatehistorybasedlocalhist}
\end{align}

\textit{Proof}
\begin{align*}
    p\fun{\state', \jointhistory' \mid \history_a} 
    &= \expectedsymbol{\state, \jointhistory_{-a} \sim p\fun{\cdot \mid \history_a}} p\fun{\state', \jointhistory' \mid \state, \tuple{\jointhistory_{-a}, \history_a}} \tag{law of total probability}\\
    &= \expectedsymbol{\state, \jointhistory_{-a} \sim p\fun{\cdot \mid \history_a}} p\fun{\state' \mid \state, \tuple{\jointhistory_{-a}, \history_a}} p\fun{\jointhistory' \mid \state', \state, \tuple{\jointhistory_{-a}, \history_a}} \tag{chain rule}\\
    &= \expectedsymbol{\state, \jointhistory_{-a} \sim p\fun{\cdot \mid \history_a}} \probtransitions\fun{\state' \mid \state, \jointpolicy\fun{\tuple{\jointhistory_{-a}, \history_a}}} p\fun{\jointhistory' \mid \state', \state, \tuple{\jointhistory_{-a}, \history_a}, \jointpolicy\fun{\tuple{\jointhistory_{-a}, \history_a}}}\\
    &= \expectedsymbol{\state, \jointhistory_{-a} \sim p\fun{\cdot \mid \history_a}} \probtransitions\fun{\state' \mid \state, \jointpolicy\fun{\tuple{\jointhistory_{-a}, \history_a}}} p\fun{\jointhistory' \mid \state', \tuple{\jointhistory_{-a}, \history_a}, \jointpolicy\fun{\tuple{\jointhistory_{-a}, \history_a}}} \tag{conditional independence of $\jointhistory'$ and $\state$ given $\state', \tuple{\jointhistory_{-a}, \history_a}, \jointpolicy\fun{\tuple{\jointhistory_{-a}, \history_a}}$}
\end{align*}

\subsection{Unbiased estimator}
\begin{theorem}
For any agent $a \in \mathcal{A}$, and any realisable local history $\history_a \in \mathcal{T}_a$, and any action $\action_a \in \actions_a$
, the Q-value proxy $\tilde{Q}_a$ is an unbiased estimator of the local Q-value $Q^{\policy_a}$
\begin{align}
    \expectedsymbol{\state, \jointhistory_{-a} \sim p\fun{\cdot \mid \history_a}} \qvaluesproxysymbol{a}{}\fun{\state, \tuple{\jointhistory_{-a}, \history_{a}}, \action_a} = \qvaluessymbol{}{\policy_a}\fun{\history_a, \action_a}
\end{align}
\end{theorem}

\textbf{Proof}\\
We fix $a \in \mathcal{A}, \action_a \in \actions_a, \history_a \in \mathcal{T}_a$

\begin{align}
    \label{eq:proof_expected_diff}
    \expected{\state, \jointhistory_{-a} \sim p\fun{\cdot \mid \history_a}}{\qvaluesproxysymbol{a}{}\fun{\state, \tuple{\jointhistory_{-a}, \history_{a}}, \action_a} - \qvaluessymbol{}{\policy_a}\fun{\history_a, \action_a}} 
    &= \expectedsymbol{\state, \jointhistory_{-a} \sim p\fun{\cdot \mid \history_a}} \left[ \valuessymbol{}{\jointpolicy}\fun{\state, \tuple{\jointhistory_{-a}, \history_{a}}} + \advantagessymbol{}{\policy_{a}}\fun{\history_a, \action_a} \right. \\
    &\quad\quad\quad\quad\quad\quad\quad\quad - \left. \left( \valuessymbol{}{\policy_{a}}\fun{\history_a} + \advantagessymbol{}{\policy_{a}}\fun{\history_a, \action_a} \right) \right] \nonumber\\
    &=  \expected{\state, \jointhistory_{-a} \sim p\fun{\cdot \mid \history_a}}{\valuessymbol{}{\jointpolicy}\fun{\state, \tuple{\jointhistory_{-a}, \history_{a}}} -  \valuessymbol{}{\policy_{a}}\fun{\history_a} } \nonumber
\end{align}

By definition we have:
\begin{align*}
    \valuessymbol{}{\jointpolicy}\fun{\state, \jointhistory} &= r\fun{\state, \jointpolicy\fun{\jointhistory}} + \gamma \expectedsymbol{\state' \sim \probtransitions\fun{\cdot \mid \state, \jointpolicy\fun{\jointhistory}}}
    \expectedsymbol{\jointhistory' \sim p\fun{\cdot \mid \state', \jointpolicy\fun{\jointhistory}, \jointhistory}} \valuessymbol{}{\jointpolicy}\fun{\state', \jointhistory'} \\
    \valuessymbol{}{\policy_a}\fun{\history_a} &= \expectedsymbol{s, \jointhistory_{-a} \sim p\fun{\cdot \mid \history_a}} 
    r\fun{\state, \jointpolicy\fun{\tuple{\jointhistory_{-a}, \history_{a}}}} + \gamma  \expectedsymbol{s, \jointhistory_{-a} \sim p\fun{\cdot \mid \history_a}} \expectedsymbol{\state' \sim \probtransitions\fun{\cdot \mid \state, \jointpolicy\fun{\tuple{\jointhistory_{-a}, \history_{a}}}}}
    \expectedsymbol{\history_a' \sim p\fun{\cdot \mid \state', \jointpolicy\fun{\tuple{\jointhistory_{-a}, \history_{a}}}, \tuple{\jointhistory_{-a}, \history_{a}}}} \valuessymbol{}{\policy_a}\fun{\history_a'}    
\end{align*}

We define $\Delta_r$ and $\Delta_p$ as follow:
\begin{align*}
    \Delta_r\fun{\state, \tuple{\jointhistory_{-a}, \history_{a}}} &= r\fun{\state, \jointpolicy\fun{\tuple{\jointhistory_{-a}, \history_{a}}}} - \expectedsymbol{\tilde{\state}, \tilde{\jointhistory}_{-a} \sim p\fun{\cdot \mid \history_a}} 
    r\fun{\tilde{\state}, \jointpolicy\fun{\tuple{\tilde{\jointhistory}_{-a}, \history_{a}}}} \\
    \Delta_p\fun{\state, \tuple{\jointhistory_{-a}, \history_{a}}} &= \expectedsymbol{\state' \sim \probtransitions\fun{\cdot \mid \state, \jointpolicy\fun{\tuple{\jointhistory_{-a}, \history_{a}}}}}
    \expectedsymbol{\jointhistory' \sim p\fun{\cdot \mid \state', \jointpolicy\fun{\tuple{\jointhistory_{-a}, \history_{a}}}, \tuple{\jointhistory_{-a}, \history_{a}}}} \valuessymbol{}{\jointpolicy}\fun{\state', \jointhistory'} \\
    &\quad -
    \expectedsymbol{\tilde{\state}, \tilde{\jointhistory}_{-a} \sim p\fun{\cdot \mid \history_a}} \expectedsymbol{\state' \sim \probtransitions\fun{\cdot \mid \tilde{\state}, \jointpolicy\fun{\tuple{\tilde{\jointhistory}_{-a}, \history_{a}}}}}
    \expectedsymbol{\history_a' \sim p\fun{\cdot \mid \tilde{\state}', \jointpolicy\fun{\tuple{\tilde{\jointhistory}_{-a}, \history_{a}}}, \tuple{\tilde{\jointhistory_{-a}}, \history_{a}}}} \valuessymbol{}{\policy_a}\fun{\history_a'}  
\end{align*}

This allows us to rewrite Eq \ref{eq:proof_expected_diff} as

\begin{align}
    \expected{\state, \jointhistory_{-a} \sim p\fun{\cdot \mid \history_a}}{\qvaluesproxysymbol{a}{}\fun{\state, \tuple{\jointhistory_{-a}, \history_{a}}, \action_a} - \qvaluessymbol{}{\policy_a}\fun{\history_a, \action_a}} &= 
    \expected{\state, \jointhistory_{-a} \sim p\fun{\cdot \mid \history_a}}{\Delta_r\fun{\state, \tuple{\jointhistory_{-a}, \history_{a}}}} 
    + \gamma \expected{\state, \jointhistory_{-a} \sim p\fun{\cdot \mid \history_a}}{\Delta_p\fun{\state, \tuple{\jointhistory_{-a}, \history_{a}}}}
    \label{eq:proof_expected_diff_delta}
\end{align}

Let's first focus on the first part of the RHS of Equation \ref{eq:proof_expected_diff}.

\begin{align*}
    \expected{\state, \jointhistory_{-a} \sim p\fun{\cdot \mid \history_a}}{\Delta_r\fun{\state, \tuple{\jointhistory_{-a}, \history_{a}}}}  
    &= \expected{\state, \jointhistory_{-a} \sim p\fun{\cdot \mid \history_a}}{r\fun{\state, \jointpolicy\fun{\tuple{\jointhistory_{-a}, \history_{a}}}} - \expectedsymbol{\tilde{\state}, \tilde{\jointhistory}_{-a} \sim p\fun{\cdot \mid \history_a}} 
    r\fun{\tilde{\state}, \jointpolicy\fun{\tuple{\tilde{\jointhistory}_{-a}, \history_{a}}}}} \\
    &= \expectedsymbol{\state, \jointhistory_{-a} \sim p\fun{\cdot \mid \history_a}}{r\fun{\state, \jointpolicy\fun{\tuple{\jointhistory_{-a}, \history_{a}}}}} - \expectedsymbol{\state, \jointhistory_{-a} \sim p\fun{\cdot \mid \history_a}}{\expectedsymbol{\tilde{\state}, \tilde{\jointhistory}_{-a} \sim p\fun{\cdot \mid \history_a}} 
    r\fun{\tilde{\state}, \jointpolicy\fun{\tuple{\tilde{\jointhistory}_{-a}, \history_{a}}}}} \tag{linearity of expectation} \\
    &= \expectedsymbol{\state, \jointhistory_{-a} \sim p\fun{\cdot \mid \history_a}}{r\fun{\state, \jointpolicy\fun{\tuple{\jointhistory_{-a}, \history_{a}}}}} - {\expectedsymbol{\tilde{\state}, \tilde{\jointhistory}_{-a} \sim p\fun{\cdot \mid \history_a}} 
    r\fun{\tilde{\state}, \jointpolicy\fun{\tuple{\tilde{\jointhistory}_{-a}, \history_{a}}}}} \tag{second part does not depend on $s, \jointhistory_{-a}$} \\
    &= 0
\end{align*}

Let's now focus on the second part of the RHS of Equation \ref{eq:proof_expected_diff}.

\begin{align*}
   \expected{\state, \jointhistory_{-a} \sim p\fun{\cdot \mid \history_a}}{\Delta_p\fun{\state, \tuple{\jointhistory_{-a}, \history_{a}}}} 
   &= \expectedsymbol{\state, \jointhistory_{-a} \sim p\fun{\cdot \mid \history_a}} \left[
   \expectedsymbol{\state' \sim \probtransitions\fun{\cdot \mid \state, \jointpolicy\fun{\tuple{\jointhistory_{-a}, \history_{a}}}}}
    \expectedsymbol{\jointhistory' \sim p\fun{\cdot \mid \state', \jointpolicy\fun{\tuple{\jointhistory_{-a}, \history_{a}}}, \tuple{\jointhistory_{-a}, \history_{a}}}} \valuessymbol{}{\jointpolicy}\fun{\state', \jointhistory'} \right. \\ 
    & \quad\quad\quad \left. - \expectedsymbol{\tilde{\state}, \tilde{\jointhistory}_{-a} \sim p\fun{\cdot \mid \history_a}} \expectedsymbol{\state' \sim \probtransitions\fun{\cdot \mid \tilde{\state}, \jointpolicy\fun{\tuple{\tilde{\jointhistory}_{-a}, \history_{a}}}}}
    \expectedsymbol{\history_a' \sim p\fun{\cdot \mid \tilde{\state}', \jointpolicy\fun{\tuple{\tilde{\jointhistory}_{-a}, \history_{a}}}, \tuple{\tilde{\jointhistory_{-a}}, \history_{a}}}} \valuessymbol{}{\policy_a}\fun{\history_a'} \right] \\
    &= \overbrace{\expectedsymbol{\state, \jointhistory_{-a} \sim p\fun{\cdot \mid \history_a}}
   \expectedsymbol{\state' \sim \probtransitions\fun{\cdot \mid \state, \jointpolicy\fun{\tuple{\jointhistory_{-a}, \history_{a}}}}}
    \expectedsymbol{\jointhistory' \sim p\fun{\cdot \mid \state', \jointpolicy\fun{\tuple{\jointhistory_{-a}, \history_{a}}}, \tuple{\jointhistory_{-a}, \history_{a}}}} \valuessymbol{}{\jointpolicy}\fun{\state', \jointhistory'}}^\text{A} \\ 
    & \quad\quad\quad  - \underbrace{\expectedsymbol{\state, \jointhistory_{-a} \sim p\fun{\cdot \mid \history_a}}\expectedsymbol{\tilde{\state}, \tilde{\jointhistory}_{-a} \sim p\fun{\cdot \mid \history_a}} \expectedsymbol{\state' \sim \probtransitions\fun{\cdot \mid \tilde{\state}, \jointpolicy\fun{\tuple{\tilde{\jointhistory}_{-a}, \history_{a}}}}}
    \expectedsymbol{\history_a' \sim p\fun{\cdot \mid \tilde{\state}', \jointpolicy\fun{\tuple{\tilde{\jointhistory}_{-a}, \history_{a}}}, \tuple{\tilde{\jointhistory_{-a}}, \history_{a}}}} \valuessymbol{}{\policy_a}\fun{\history_a'}}_\text{B} \tag{linearity of expectation} 
\end{align*}

\begin{align*}
   \text{A} 
    &= \expectedsymbol{\state, \jointhistory_{-a} \sim p\fun{\cdot \mid \history_a}}
   \expectedsymbol{\state' \sim \probtransitions\fun{\cdot \mid \state, \jointpolicy\fun{\tuple{\jointhistory_{-a}, \history_{a}}}}}
    \expectedsymbol{\jointhistory' \sim p\fun{\cdot \mid \state', \jointpolicy\fun{\tuple{\jointhistory_{-a}, \history_{a}}}, \tuple{\jointhistory_{-a}, \history_{a}}}} \valuessymbol{}{\jointpolicy}\fun{\state', \jointhistory'}\\
    &= \expectedsymbol{\state, \jointhistory_{-a} \sim p\fun{\cdot \mid \history_a}}
    \int_{\state'} \probtransitions\fun{\state' \mid \state, \jointpolicy\fun{\tuple{\jointhistory_{-a}, \history_{a}}}}
    \int_{\jointhistory'} p\fun{\jointhistory' \mid \state', \jointpolicy\fun{\tuple{\jointhistory_{-a}, \history_{a}}}, \tuple{\jointhistory_{-a}, \history_{a}}} \valuessymbol{}{\jointpolicy}\fun{\state', \jointhistory'} \,\text{d}\state'\,\text{d}\jointhistory' 
    \tag{definition of expectation} \\
    &= \int_{\state'} \int_{\jointhistory'} \expectedsymbol{\state, \jointhistory_{-a} \sim p\fun{\cdot \mid \history_a}}
     \probtransitions\fun{\state' \mid \state, \jointpolicy\fun{\tuple{\jointhistory_{-a}, \history_{a}}}}
    p\fun{\jointhistory' \mid \state', \jointpolicy\fun{\tuple{\jointhistory_{-a}, \history_{a}}}, \tuple{\jointhistory_{-a}, \history_{a}}} \valuessymbol{}{\jointpolicy}\fun{\state', \jointhistory'} \,\text{d}\state'\,\text{d}\jointhistory' 
    \tag{linearity} \\
    &= \int_{\state'} \int_{\jointhistory'} p\fun{\state', \jointhistory' \mid \history_a} \valuessymbol{}{\jointpolicy}\fun{\state', \jointhistory'} \,\text{d}\state'\,\text{d}\jointhistory' 
    \tag{see Eq. \ref{eq:nextstatehistorybasedlocalhist}} \\
    &= \expectedsymbol{\state', \jointhistory' \sim p\fun{\cdot \mid \history_a}} \valuessymbol{}{\jointpolicy}\fun{\state', \jointhistory'}
    \tag{definition of expectation}
\end{align*}

\begin{align*}
    \text{B}
    &= \expectedsymbol{\state, \jointhistory_{-a} \sim p\fun{\cdot \mid \history_a}}\expectedsymbol{\tilde{\state}, \tilde{\jointhistory}_{-a} \sim p\fun{\cdot \mid \history_a}} \expectedsymbol{\state' \sim \probtransitions\fun{\cdot \mid \tilde{\state}, \jointpolicy\fun{\tuple{\tilde{\jointhistory}_{-a}, \history_{a}}}}}
    \expectedsymbol{\history_a' \sim p\fun{\cdot \mid \tilde{\state}', \jointpolicy\fun{\tuple{\tilde{\jointhistory}_{-a}, \history_{a}}}, \tuple{\tilde{\jointhistory_{-a}}, \history_{a}}}} \valuessymbol{}{\policy_a}\fun{\history_a'} \\
    &= \expectedsymbol{\tilde{\state}, \tilde{\jointhistory}_{-a} \sim p\fun{\cdot \mid \history_a}} \expectedsymbol{\state' \sim \probtransitions\fun{\cdot \mid \tilde{\state}, \jointpolicy\fun{\tuple{\tilde{\jointhistory}_{-a}, \history_{a}}}}}
    \expectedsymbol{\history_a' \sim p\fun{\cdot \mid \tilde{\state}', \jointpolicy\fun{\tuple{\tilde{\jointhistory}_{-a}, \history_{a}}}, \tuple{\tilde{\jointhistory_{-a}}, \history_{a}}}} \valuessymbol{}{\policy_a}\fun{\history_a'} \tag{does not depend on $\state, \jointhistory_{-a}$} \\
    &= \expectedsymbol{\tilde{\state}, \tilde{\jointhistory}_{-a} \sim p\fun{\cdot \mid \history_a}} \expectedsymbol{\state' \sim \probtransitions\fun{\cdot \mid \tilde{\state}, \jointpolicy\fun{\tuple{\tilde{\jointhistory}_{-a}, \history_{a}}}}}
    \int_{\history_a'} p\fun{\history_a' \mid \tilde{\state}', \jointpolicy\fun{\tuple{\tilde{\jointhistory}_{-a}, \history_{a}}}, \tuple{\tilde{\jointhistory_{-a}}, \history_{a}}} \valuessymbol{}{\policy_a}\fun{\history_a'} \,\text{d}\history_a'
    \tag{definition of expectation} \\
    &= \int_{\history_a'} \expectedsymbol{\tilde{\state}, \tilde{\jointhistory}_{-a} \sim p\fun{\cdot \mid \history_a}} \expectedsymbol{\state' \sim \probtransitions\fun{\cdot \mid \tilde{\state}, \jointpolicy\fun{\tuple{\tilde{\jointhistory}_{-a}, \history_{a}}}}}
    p\fun{\history_a' \mid \tilde{\state}', \jointpolicy\fun{\tuple{\tilde{\jointhistory}_{-a}, \history_{a}}}, \tuple{\tilde{\jointhistory_{-a}}, \history_{a}}} \valuessymbol{}{\policy_a}\fun{\history_a'} \,\text{d}\history_a'
    \tag{linearity} \\
    &= \int_{\history_a'} 
    p\fun{\history_a' \mid \history_{a}} \valuessymbol{}{\policy_a}\fun{\history_a'} \,\text{d}\history_a'
    \tag{see Eq. \ref{eq:nextlocalhistorybasedlocalhistory}} \\
    &= \expectedsymbol{\history_a' \sim p\fun{\cdot \mid \history_{a}}} 
    \valuessymbol{}{\policy_a}\fun{\history_a'}
    \tag{definition of expectation} \\
\end{align*}

By using the value of A and B we get:

\begin{align*}
    \expected{\state, \jointhistory_{-a} \sim p\fun{\cdot \mid \history_a}}{\Delta_p\fun{\state, \tuple{\jointhistory_{-a}, \history_{a}}}} 
    &= \expectedsymbol{\state', \jointhistory' \sim p\fun{\cdot \mid \history_a}} \valuessymbol{}{\jointpolicy}\fun{\state', \jointhistory'} - \expectedsymbol{\history_a' \sim p\fun{\cdot \mid \history_{a}}} 
    \valuessymbol{}{\policy_a}\fun{\history_a'} \\
    &= \expectedsymbol{\history_a' \sim p\fun{\cdot \mid \history_a}} \expectedsymbol{\state', \jointhistory_{-a}' \sim p\fun{\cdot \mid \history_a, \history_a'}} \valuessymbol{}{\jointpolicy}\fun{\state', \tuple{\jointhistory_{-a}', \history_a'}}  - \expectedsymbol{\history_a' \sim p\fun{\cdot \mid \history_{a}}} 
    \valuessymbol{}{\policy_a}\fun{\history_a'}
    \tag{chain rule} \\
    &= \expectedsymbol{\history_a' \sim p\fun{\cdot \mid \history_a}} \expectedsymbol{\state', \jointhistory_{-a}' \sim p\fun{\cdot \mid \history_a'}} \valuessymbol{}{\jointpolicy}\fun{\state', \tuple{\jointhistory_{-a}', \history_a'}}  - \expectedsymbol{\history_a' \sim p\fun{\cdot \mid \history_{a}}} 
    \valuessymbol{}{\policy_a}\fun{\history_a'}
    \tag{$\history_a'$ contains $\history_a$} \\
    &= \expectedsymbol{\history_a' \sim p\fun{\cdot \mid \history_{a}}}\expected{\state', \jointhistory_{-a}' \sim p\fun{\cdot \mid \history_a'}}{
    \valuessymbol{}{\jointpolicy}\fun{\state', \tuple{\jointhistory_{-a}', \history_a'}}   -  
    \valuessymbol{}{\policy_a}\fun{\history_a'}}
    \tag{linearity} \\
\end{align*}

Therefore we obtain:
\begin{align}
     \expected{\state, \jointhistory_{-a} \sim p\fun{\cdot \mid \history_a}}{\valuessymbol{}{\jointpolicy}\fun{\state, \tuple{\jointhistory_{-a}, \history_{a}}} -  \valuessymbol{}{\policy_{a}}\fun{\history_a} } 
     &= \gamma \expectedsymbol{\history_a' \sim p\fun{\cdot \mid \history_{a}}}\expected{\state', \jointhistory_{-a}' \sim p\fun{\cdot \mid \history_a'}}{
    \valuessymbol{}{\jointpolicy}\fun{\state', \tuple{\jointhistory_{-a}', \history_a'}}   -  
    \valuessymbol{}{\policy_a}\fun{\history_a'}}
    \label{eq:value_gamma}
\end{align}

By applying recursively $n$ times Equation \ref{eq:value_gamma} we obtain: 
\begin{align}
     \expected{\state, \jointhistory_{-a} \sim p\fun{\cdot \mid \history_a}}{\valuessymbol{}{\jointpolicy}\fun{\state, \tuple{\jointhistory_{-a}, \history_{a}}} -  \valuessymbol{}{\policy_{a}}\fun{\history_a} } & \nonumber\\
     &\hspace{-5em}= \gamma^n \expectedsymbol{\history_a^{1} \sim p\fun{\cdot \mid \history_{a}}} \expectedsymbol{\history_a^{2} \sim p\fun{\cdot \mid \history_{a}^{1}}} \text{...} \expectedsymbol{\history_a^{n} \sim p\fun{\cdot \mid \history_{a}^{n-1}}} \expected{\state^{n}, \jointhistory_{-a}^{n} \sim p\fun{\cdot \mid \history_a^{n}}}{
    \valuessymbol{}{\jointpolicy}\fun{\state^{n}, \tuple{\jointhistory_{-a}^{n}, \history_a^{n}}}   -  
    \valuessymbol{}{\policy_a}\fun{\history_a^{n}}}
    \label{eq:value_gamma_req}
\end{align}

We then define $R_{\max} = \max_{\state \in \states} \max_{\jointaction \in \bm{\actions}} \abs{R\fun{\state, \jointaction}}$. This allows us to bound the difference between the centralized value and the local value:
\begin{align*}
    \forall \state \in \states, \jointhistory \in \bm{\mathcal{T}},\quad \abs{
    \valuessymbol{}{\jointpolicy}\fun{\state, \tuple{\jointhistory_{-a}, \history_a}}   -  
    \valuessymbol{}{\policy_a}\fun{\history_a}
    } &\leq \abs{\valuessymbol{}{\jointpolicy}\fun{\state, \tuple{\jointhistory_{-a}, \history_a}}} + \abs{\valuessymbol{}{\policy_a}\fun{\history_a}} \tag{triangular inequality} \\
    &\leq \frac{R_{\max}}{1 - \gamma} + \frac{R_{\max}}{1 - \gamma} \tag{upper-bound on the value}\\
    &\leq \frac{2R_{\max}}{1 - \gamma}
\end{align*}

This allows us to bound the LHS of Equation \ref{eq:value_gamma_req}:
\begin{align*}
     \abs{\expected{\state, \jointhistory_{-a} \sim p\fun{\cdot \mid \history_a}}{\valuessymbol{}{\jointpolicy}\fun{\state, \tuple{\jointhistory_{-a}, \history_{a}}} -  \valuessymbol{}{\policy_{a}}\fun{\history_a} } }& \\
     &\hspace{-5em}= \abs{\gamma^n \expectedsymbol{\history_a^{1} \sim p\fun{\cdot \mid \history_{a}}} \expectedsymbol{\history_a^{2} \sim p\fun{\cdot \mid \history_{a}^{1}}} \text{...} \expectedsymbol{\history_a^{n} \sim p\fun{\cdot \mid \history_{a}^{n-1}}} \expected{\state^{n}, \jointhistory_{-a}^{n} \sim p\fun{\cdot \mid \history_a^{n}}}{
    \valuessymbol{}{\jointpolicy}\fun{\state^{n}, \tuple{\jointhistory_{-a}^{n}, \history_a^{n}}}   -  
    \valuessymbol{}{\policy_a}\fun{\history_a^{n}}}}\\
    &\hspace{-5em}\leq \gamma^n \expectedsymbol{\history_a^{1} \sim p\fun{\cdot \mid \history_{a}}} \expectedsymbol{\history_a^{2} \sim p\fun{\cdot \mid \history_{a}^{1}}} \text{...} \expectedsymbol{\history_a^{n} \sim p\fun{\cdot \mid \history_{a}^{n-1}}} \expected{\state^{n}, \jointhistory_{-a}^{n} \sim p\fun{\cdot \mid \history_a^{n}}}{\abs{
    \valuessymbol{}{\jointpolicy}\fun{\state^{n}, \tuple{\jointhistory_{-a}^{n}, \history_a^{n}}}   -  
    \valuessymbol{}{\policy_a}\fun{\history_a^{n}}}} \tag{Jensen inequality} \\
    &\hspace{-5em}\leq \gamma^n \expectedsymbol{\history_a^{1} \sim p\fun{\cdot \mid \history_{a}}} \expectedsymbol{\history_a^{2} \sim p\fun{\cdot \mid \history_{a}^{1}}} \text{...} \expectedsymbol{\history_a^{n} \sim p\fun{\cdot \mid \history_{a}^{n-1}}} \expected{\state^{n}, \jointhistory_{-a}^{n} \sim p\fun{\cdot \mid \history_a^{n}}}{  \frac{2R_{\max}}{1 - \gamma}} \tag{see above} \\
    &\hspace{-5em}\leq \gamma^n \frac{2R_{\max}}{1 - \gamma} 
\end{align*}

As $\gamma \in ]0, 1[$, when $n \rightarrow +\inf$ we obtain:
\begin{align*}
     \abs{\expected{\state, \jointhistory_{-a} \sim p\fun{\cdot \mid \history_a}}{\valuessymbol{}{\jointpolicy}\fun{\state, \tuple{\jointhistory_{-a}, \history_{a}}} -  \valuessymbol{}{\policy_{a}}\fun{\history_a} } } \leq 0 
\end{align*}

And finally, using Eq. \ref{eq:proof_expected_diff}:
\begin{align*}
    \expected{\state, \jointhistory_{-a} \sim p\fun{\cdot \mid \history_a}}{\qvaluesproxysymbol{a}{}\fun{\state, \tuple{\jointhistory_{-a}, \history_{a}}, \action_a} - \qvaluessymbol{}{\policy_a}\fun{\history_a, \action_a}} = 0
\end{align*}

\section{Additional Experiment}
\label{sec:mpe}
\begin{figure}
    \centering
    \includegraphics[width=0.9\linewidth]{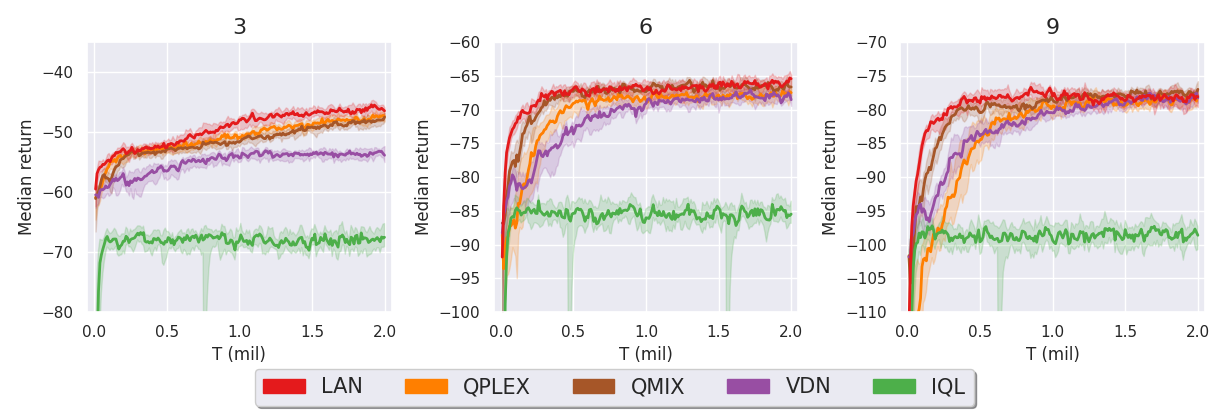}
    \caption{Median return during learning on the simple spread environment of MPE with 3, 6 and 9 agent. Each algorithm is run on 10 different seeds. We train the agents on 2 million steps and plot the median, 1st and 3rd quantiles.}
    \label{fig:mpe}
\end{figure}

We conducted an evaluation of LAN and the selected baseline algorithms within a modified version of the \textit{simple spread} environment from the Multi-Agent Particle Environment suite (MPE) \cite{Lowe2017Multi-agentEnvironments}. In the original environment, three agents are tasked to spread efficiently across three landmarks while avoiding collisions with one another. The reward structure combined two components: a) the cumulative negative distance between each landmark and its closest agent; b) penalties for collisions between agents. Both agents and landmarks are randomly spawned on the map at the beginning of an episode. Notably, we introduced partial observability into the environment, restricting agents to observe only those agents and landmarks within a fixed radius. Additionally, modifications were made to the environment, allowing for any number of agents while maintaining a constant ratio between the environment size and the agent count.

Figure [\ref{fig:mpe} depicts the median return obtained by LAN and the baseline algorithms with 3, 6, and 9 agents. The results are averaged over 10 runs. The hyper-parameters from SMAC were adopted without further tuning. The results consistently demonstrate LAN's accelerated learning compared to other algorithms in all three instances. With 3 agents, both QPLEX and QMIX exhibit slightly inferior performance relative to LAN. In contrast, VDN significantly underperforms in comparison to LAN, while IQL appears to struggle in learning. With 6 and 9 agents, the learning curves of LAN, QPLEX, QMIX, and VDN align closely, eventually reaching similar performance levels. However we note that LAN consistently achieves quicker convergence. IQL fails to learn a good policy in all the instances.

\end{document}